\documentclass[11pt]{article}

\usepackage[final]{acl}

\usepackage{times}
\usepackage{latexsym}

\usepackage[T1]{fontenc}
\usepackage[utf8]{inputenc}

\usepackage{microtype}

\usepackage{inconsolata}

\usepackage{amsmath}
\usepackage{amssymb}
\usepackage{braket}
\usepackage{amsthm}
\usepackage{multirow}

\usepackage{graphicx}
\usepackage{mathtools}
\usepackage{xcolor}
\usepackage{listings}
\usepackage{tabularx}
\usepackage{booktabs}

\definecolor{KBase}{HTML}{0072B2}      % blue
\definecolor{KExt}{HTML}{D55E00}       % orange
\definecolor{KMand}{HTML}{6A3D9A}      % purple
\definecolor{KKron}{HTML}{009E73}    % green

\newcommand{\Kbase}[1]{\textcolor{KBase}{#1}}
\newcommand{\Kext}[1]{\textcolor{KExt}{#1}}

\lstdefinestyle{python}{
    language=Python,
    basicstyle=\ttfamily\footnotesize,
    breaklines=true,
    columns=fullflexible,
    keepspaces=true,
    showstringspaces=false,
    frame=single,
    numbers=left,
    numberstyle=\tiny,
    commentstyle=\ttfamily\upshape,
    tabsize=4
}

\lstdefinestyle{aclpython}{
    style=python,
    basicstyle=\ttfamily\scriptsize
}
\title{Kronecker Factorization Improves Efficiency and Interpretability of Sparse Autoencoders}

\author{
 \textbf{Vadim Kurochkin},
 \textbf{Yaroslav Aksenov}\textsuperscript{*},
\\
 \textbf{Daniil Laptev},
 \textbf{Daniil Gavrilov},
 \textbf{Nikita Balagansky}
\\
 T-Tech \\
 \small{
   \textsuperscript{*}\textbf{Correspondence:} Yaroslav Aksenov,
   \href{mailto:y.o.aksenov@t-tech.dev}{\texttt{y.o.aksenov@t-tech.dev}}
 }
}

\begin{document}

\maketitle
\begin{abstract}

Sparse Autoencoders (SAEs) decompose language-model activations into sparse, interpretable features, but standard encoders usually treat the latent dictionary as a flat set of independent coordinates, leaving hierarchy and feature interactions to emerge only implicitly. We propose \textbf{KronSAE}, a design that factorizes the latent space into heads and forms post-latent features as pairwise compositions of lower-dimensional pre-latents using \(\operatorname{mAND}\), a differentiable AND-like interaction. This imposes a compositional co-activation prior while remaining compatible with standard SAE objectives and variants such as TopK, Matryoshka, and Switch SAEs. KronSAE matches strong baselines on the EV--FLOPs frontier, improves interpretability of the latents, better captures the underlying correlated feature structure, and reduces encoder computational cost as an additional benefit. Code is available at \url{https://github.com/corl-team/kronsae}.

\end{abstract}

\section{Introduction}

\begin{figure}[ht]
    \centering
    \includegraphics[width=0.85\linewidth]{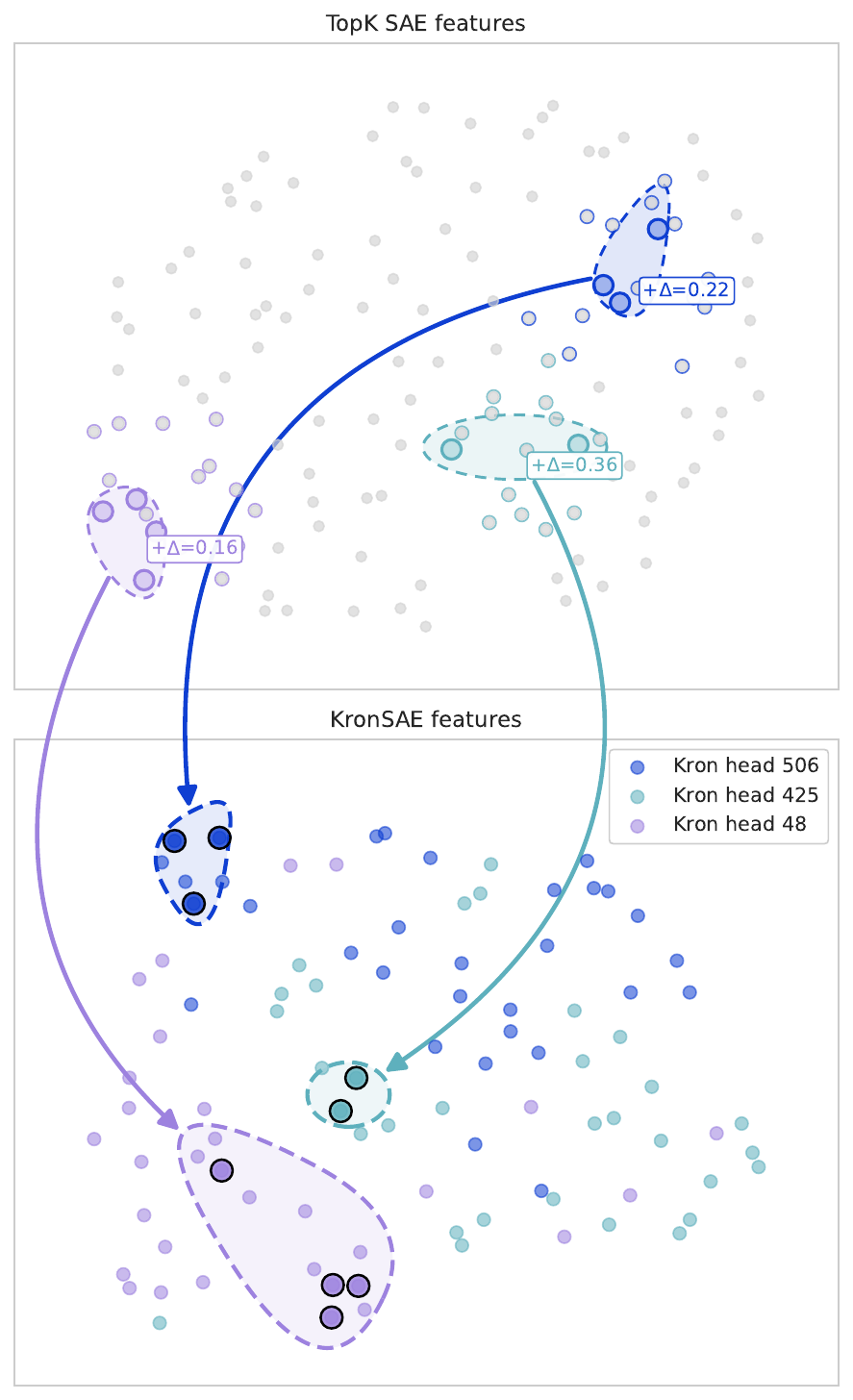}
    \caption{
    UMAP projection illustrating the relationship between TopK SAE features and KronSAE heads. Correlated TopK features tend to map to nearby KronSAE features within the same head. Selected features correlate more strongly than their nearest neighbors; $\Delta$ denotes the difference between correlations.
    }
    \label{fig:feature_flow}
\end{figure}

Interpreting the hidden states of large language models remains a central challenge for transparency and controllability in AI systems \citep{bricken2023monosemanticity,Heap2025}. Sparse autoencoders (SAEs) have become a standard tool for this problem: they decompose dense activations into sparse latent features, enabling circuit-level analysis \citep{marks2025sparse} and concept discovery in model internals \citep{bricken2023monosemanticity,Cunningham2024,gao2025scaling}. Most modern SAEs use an encoder to produce feature activations and a decoder as a learned dictionary of feature atoms. This design is effective, but it usually treats the latent dictionary as a flat collection of independently encoded coordinates. As a result, higher-level structure among features, such as hierarchy, compositionality, or systematic co-activation, is not built into the architecture and must either emerge implicitly during training or be recovered after training through separate analysis.

 \begin{figure*}
    \centering
    \includegraphics[width=\linewidth]{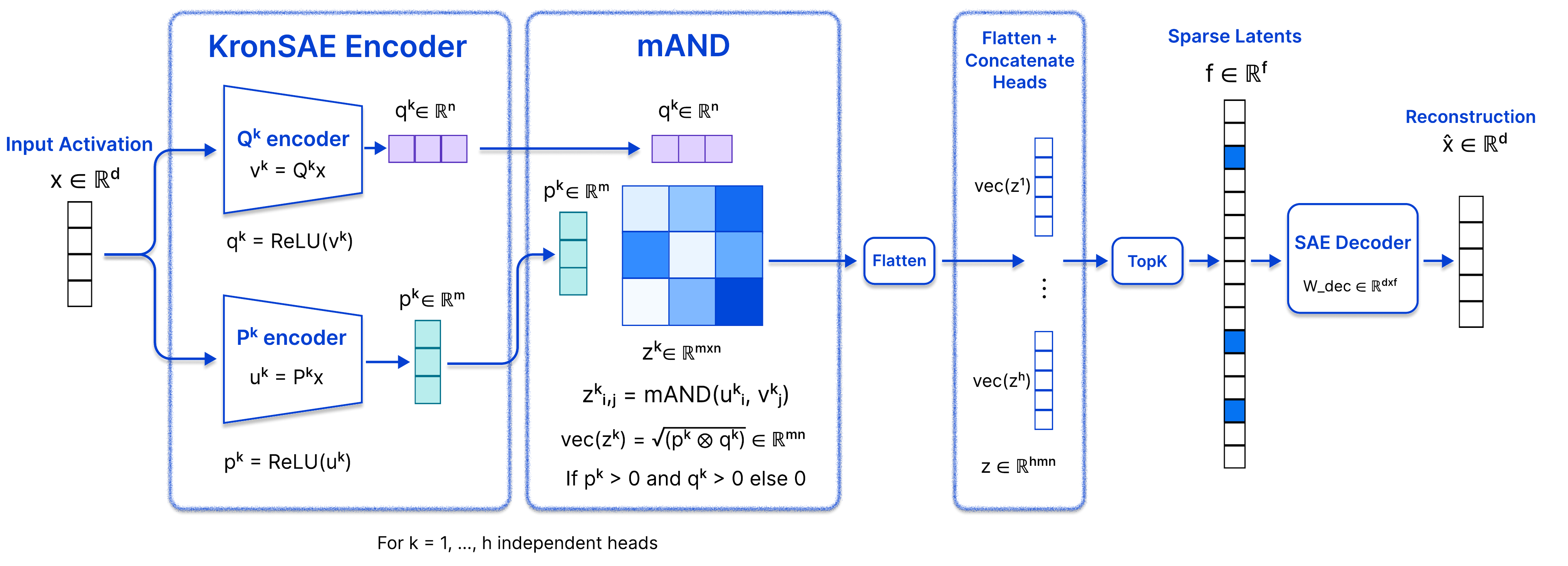}
    \caption{Overview of KronSAE. We project $x \in \mathbb{R}^d$ into head-specific base and extension pre-latents, then combine each pair with $\operatorname{mAND}$ to produce sparse post-latents.}
    \label{fig:overview}
\end{figure*}

This flat-latent default is not obviously aligned with the structure of language-model activations. Natural language data is highly structured: words and concepts have recurring statistical relationships, and transformer hidden states inherit regularities from those relationships. Recent work argues that even low-order language statistics, especially pairwise co-occurrence, can induce stable geometric structure in learned representations \citep{karkada2026symmetrylanguagestatisticsshapes}. This suggests a corresponding design principle for SAEs: if correlations and co-activations in hidden states reflect recurring linguistic and conceptual structure, then the encoder need not remain fully unconstrained. Instead, an SAE can introduce an inductive bias that makes structured feature compositions easier to learn.

Recent SAE architectures have begun to make latent organization explicit. Switch SAEs introduce a routed expert decomposition of the encoder/decoder computation \citep{mudide2025efficient}, while Matryoshka SAEs impose a nested-dictionary hierarchy over features \citep{bussmann2025learning}, yielding a more structured latent space and improved interpretability results. However, a complementary direction remains underexplored: imposing hierarchy by changing the default structure of the encoding process itself. This can provide more flexible feature modeling and make structured feature interactions easier to learn.

\textbf{We propose KronSAE, an encoder-side hierarchical modification to existing SAEs.} KronSAE replaces a dense projection to $F$ independent latents with projections to lower-dimensional pre-latents grouped into heads, and forms post-latent features through pairwise $\operatorname{mAND}$ interactions. This gives each post-latent an explicit compositional structure: it activates only when both of its parent pre-latents are active, creating an architectural prior toward co-activation and feature interaction. Figure~\ref{fig:feature_flow} gives an example of this structure. Correlated TopK SAE features tend to map to nearby KronSAE features within the same head, suggesting that some feature relationships are better described as structured co-activation patterns than as independent flat coordinates. This motivates an encoder architecture that makes such shared structure explicit rather than leaving it to emerge only implicitly during training.

The same factorization also yields a computational side benefit. Since KronSAE replaces a dense projection to $F$ independent latents with projections to $h(m+n)$ pre-latents that generate $F=h m n$ post-latents, it reduces encoder parameters and per-token FLOPs. In this paper, however, we treat this reduction as an additional benefit. Our main comparisons use an \textbf{equal token budget}, and report FLOP and parameter savings as consequences of the hierarchical encoder.
 
 This work makes four primary contributions:
 \begin{enumerate}
    \item We identify encoder-side hierarchy as a missing design axis for sparse autoencoders, motivated by structural inductive biases in feature organization in SAEs.
    \item We introduce \textbf{KronSAE}, a Kronecker-factorized SAE encoder modification equipped with the $\operatorname{mAND}$ interaction that forms post-latent features from reusable pre-latent factors.
    \item We show that KronSAE improves SAE interpretability by decreasing feature absorption and yielding more specific latents.
    \item We demonstrate an effective reconstruction--compute trade-off: reducing encoder FLOPs by more than \textbf{40\%} incurs less than a 1\% EV drop.
 \end{enumerate}

\section{Related Work}\label{sec:related_work}

\paragraph{Sparse Autoencoders.}
Early work demonstrated that SAEs can uncover human-interpretable directions in deep models \citep{Cunningham2024,templeton2024scaling}, but extending them to the large dictionary size $F$ for modern language models with large hidden dimension $d$ is expensive: each forward pass still requires a dense $\mathcal{O}(F\,d)$ encoder projection. Most prior optimization efforts focus on the decoder side: for example, \citet{gao2025scaling} introduce a fused sparse--dense \textsc{TopK} kernel that reduces wall-clock time and memory traffic, while \citet{DBLP:conf/nips/RajamanoharanCS24} decouple activation selection from magnitude prediction to improve the sparsity--MSE trade-off. Separately, \citet{rajamanoharan2024jumpingahead} propose \textsc{JumpReLU}, a nonlinearity designed to mitigate shrinkage of large activations in sparse decoders. \citet{oneill2025compute} argue that classical SAE encoders are not expressive enough and that more complex encoders can yield more faithful results without compromising interpretability.

\paragraph{Conditional Computation.}
These schemes use mixture-of-experts routing~\citep{Shazeer2017} to avoid instantiating a full dictionary per token, sending each input to a subset of expert SAEs through a lightweight gating network~\citep{mudide2025efficient}. However, they still incur a dense encoder projection within each selected expert.

\paragraph{Factorizations and Logical Activation Functions.}
Tensor product representations \citep{smolensky1990tensorproductrepresentations} have been used to represent compositional structure in dense embeddings, closely resembling our idea of AND-like compositions. Recent studies have extended tensor product representations to transformer hidden states and in-context learning~\citep{soulos-etal-2020-discovering,smolensky2024mechanismssymbolprocessingincontext}. Separately, tensor-factorization methods have been used to compress large weight matrices in language models \citep{edalati2021kroneckerdecompositiongptcompression,wang2023learning}. In parallel, differentiable logic activations have been introduced to approximate Boolean operators smoothly~\citep{Lowe2021}. Our method synthesizes these lines of work: we embed a differentiable AND-like gate into an efficient Kronecker-factorized encoder to build compositional features while preserving end-to-end differentiability.

\section{Method}\label{sec:method}
\paragraph{Preliminaries.}\label{sec:method:prelim}
Let $\mathbf{x}\in\mathbb{R}^{d}$ denote an activation vector drawn from a pretrained transformer. A conventional $\operatorname{TopK}$ SAE \citep{gao2025scaling} produces a reconstruction $\hat{\mathbf{x}}$ of $\mathbf{x}$ via
\begin{equation}
\begin{gathered}
\mathbf{f}=\operatorname{TopK}\bigl(W_{\text{enc}}\mathbf{x} + \mathbf{b_{enc}}\bigr), \\
    \hat{\mathbf{x}}=W_{\text{dec}}\mathbf{f} + \mathbf{b_{dec}},
    \label{eq:topk}
\end{gathered}
\end{equation}
where $W_{\text{enc}}\in\mathbb{R}^{F\times d}$ and $W_{\text{dec}}\in\mathbb{R}^{d\times F}$ are dense matrices and $\mathbf{f} \in \mathbb{R}^F$ is a sparse vector retaining only the $K$ largest activations. The encoder cost therefore scales as $\mathcal{O}(Fd)$ per token.

\paragraph{KronSAE.}\label{sec:method:kron} 
We decompose the latent space into $h$ independent heads, and each head $k$ is parameterised by the composition of two thin matrices
$\Kbase{P^k\in\mathbb{R}^{m\times d}}$ (composition \Kbase{base}) and $\Kext{Q^k\in\mathbb{R}^{n\times d}}$ (composition \Kext{extension}), with dimensions $m \le n\ll d$ and $F=h\,m\,n$. The \emph{pre-latents}
\begin{equation}
    \Kbase{\mathbf{p}^k}=\operatorname{ReLU}(\Kbase{\mathbf{u}^k}) \quad \text{and} \quad
    \Kext{\mathbf{q}^k}=\operatorname{ReLU}(\Kext{\mathbf{v}^k}),
    \label{eq:prelatents}
\end{equation}
where $\Kbase{\mathbf{u}^k}=\Kbase{P^k} \mathbf{x}$ and $\Kext{\mathbf{v}^k}=\Kext{Q^k} \mathbf{x}$. These pre-latents are combined through an element-wise interaction kernel independently in each head:
\begin{equation}
\begin{gathered}
    z^k_{i,j}:=
    \operatorname{mAND}(\Kbase{u^k_{i}},\Kext{v^k_{j}}) :=
    \sqrt{\Kbase{u^k_i} \Kext{v^k_j}} \\ 
    \text{if } \Kbase{u^k_{i}}>0\text{ and }\Kext{v^k_{j}}>0, \text{ otherwise } 0,
\end{gathered}
\label{eq:mand}
\end{equation}
where $\mathbf{z}^k\in\mathbb{R}^{m\times n}$. It is then flattened in row-major order to a vector equivalent to the element-wise square root of the Kronecker product of $\Kbase{\mathbf{p}^k}$ and $\Kext{\mathbf{q}^k}$. Assuming that $\operatorname{vec}(\cdot)$ uses row-major order,
\begin{equation}
    \operatorname{vec}(\mathbf{z}^k) = \sqrt{\operatorname{vec} \left( \Kbase{\mathbf{p}^k} \Kext{(\mathbf{q}^k)^\top} \right)} = \sqrt{\Kbase{\mathbf{p}^k} \otimes \Kext{\mathbf{q}^k}}.
\end{equation}

Concatenating heads and applying $\operatorname{TopK}$ yields \emph{post-latents} $\mathbf{f}\in\mathbb{R}^{F}$. Activation events follow an AND-like rule: before TopK selection, a post-latent has support only where both parents are active; global TopK selection further restricts the final support. This architectural co-activation prior induces specific semantic structure, as we empirically verify in Section~\ref{sec:featureAnalysis}. The square root in Equation~\ref{eq:mand} keeps output magnitudes comparable. In practice, we add a small $\epsilon$ before taking the square root for numerical stability. Although this introduces slight noise into the exact AND rule, its effect is negligible after TopK selection and vanishes as $\epsilon \to 0$.

The KronSAE approach is broadly compatible with sparse decoder kernels, MoE-style routing, and different loss functions and activations, as we show in the following sections. See Appendix~\ref{sec:logical_operator} for additional discussion of the method's semantics and Appendix~\ref{sec:computational_cost} for computational costs.

\begin{figure*}[t!]
    \centering
    \includegraphics[width=\linewidth]{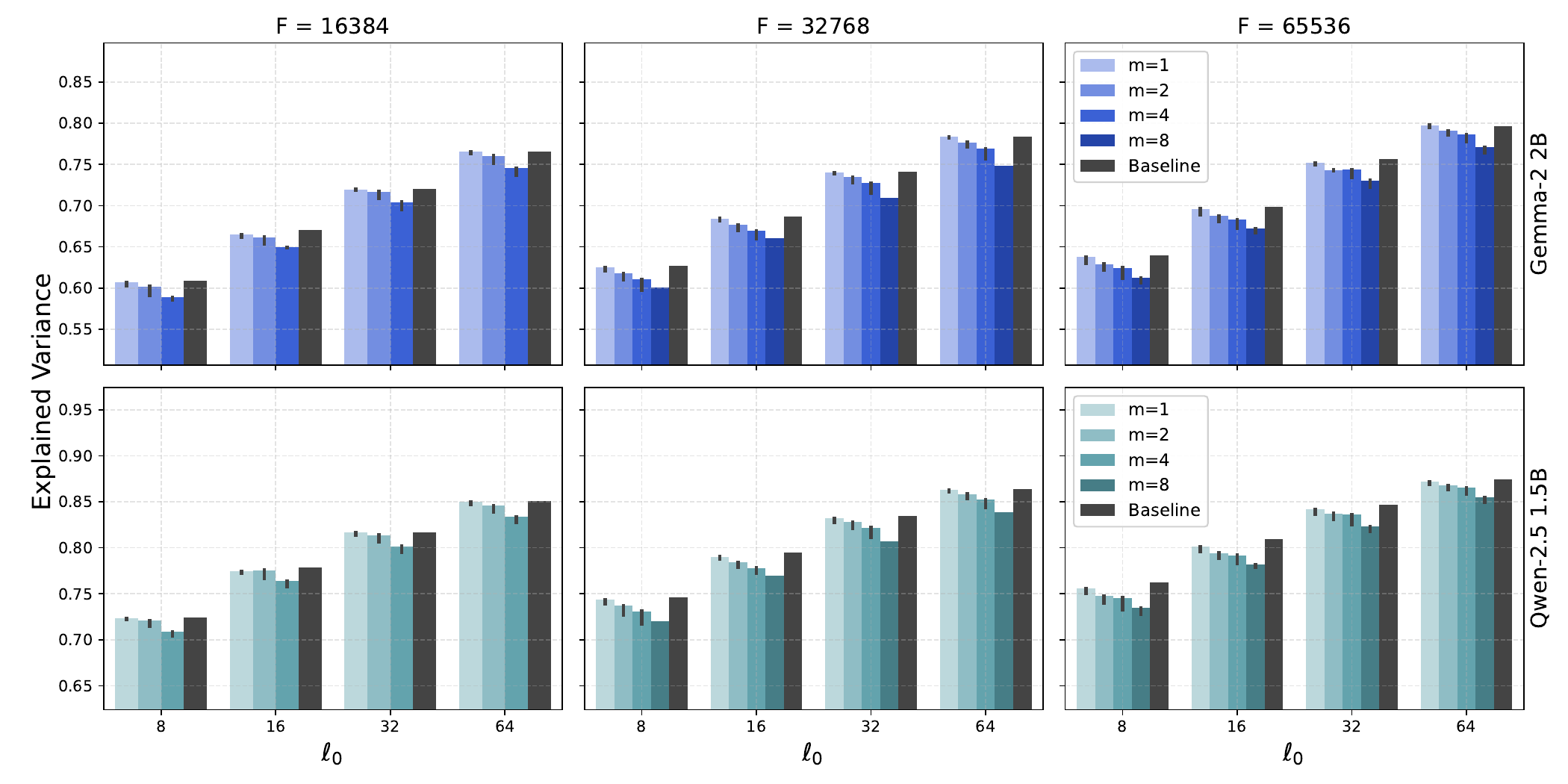}
    \caption{Maximum performance for KronSAE and TopK SAE on Qwen2.5-1.5B and Gemma-2-2B at different dictionary sizes $F$ after training on 250M tokens. KronSAE uses fewer encoder parameters while remaining on par with the baseline in EV; the gap narrows at larger dictionary sizes on layer 12 of both models.}
    \label{fig:maximum_performance_qwen_expanded}
\end{figure*}

\section{Experiments} \label{sec:ablations}

In this section, we examine hyperparameter choices, the resulting trade-offs, and KronSAE's compatibility with existing SAE variants.

\paragraph{Experimental setup.} 
We train SAEs on the residual streams of the Qwen2.5-1.5B-Base~\citep{Yang2024} and Gemma-2-2B~\citep{gemmateam2024gemma2improvingopen} language models. Activations are collected from FineWeb-Edu, a filtered subset of educational web pages from the FineWeb corpus~\citep{penedo2024finewebdatasetsdecantingweb}. We measure reconstruction quality using explained variance (EV),
\begin{equation}
    \mathrm{EV} = 1 - \frac{\operatorname{Var}(\mathbf{x} - \hat{\mathbf{x}})}{\operatorname{Var}(\mathbf{x})},
\end{equation}
where $1.0$ is optimal, and use an automated interpretability pipeline \citep{bills2023llmcanexplain,paulo2024automatically} to measure the interpretability of learned latents (see Appendix~\ref{sec:featureAnalysisDetails}).

\paragraph{Reconstruction fidelity.}
We sweep over $m \in \{1, 2, 4, 8\}$, $h \in \{512, 1024, 2048, 4096\}$, $n=F/(hm)$, $\ell_0 \in \{8, 16, 32, 64\}$, and $F \in \{2^{14}, 2^{15}, 2^{16}\}$ at early, middle, and late layers. We evaluate KronSAE against TopK SAE using a budget of 250M tokens. For most setups with fixed $m$, we report frontier performance across $n$ and $h$, together with dispersion over all configurations. As Figure~\ref{fig:maximum_performance_qwen_expanded} shows, under an equal token budget KronSAE achieves EV comparable to TopK SAE with fewer encoder FLOPs. The conclusions for layers 6 and 20 are consistent with those for layer 12 (Appendix~\ref{appendix:different_layers_ablation}). We also report MSE for all layers in Appendix~\ref{sec:mse_performance}.

\begin{figure}[ht]
  \includegraphics[width=\columnwidth]{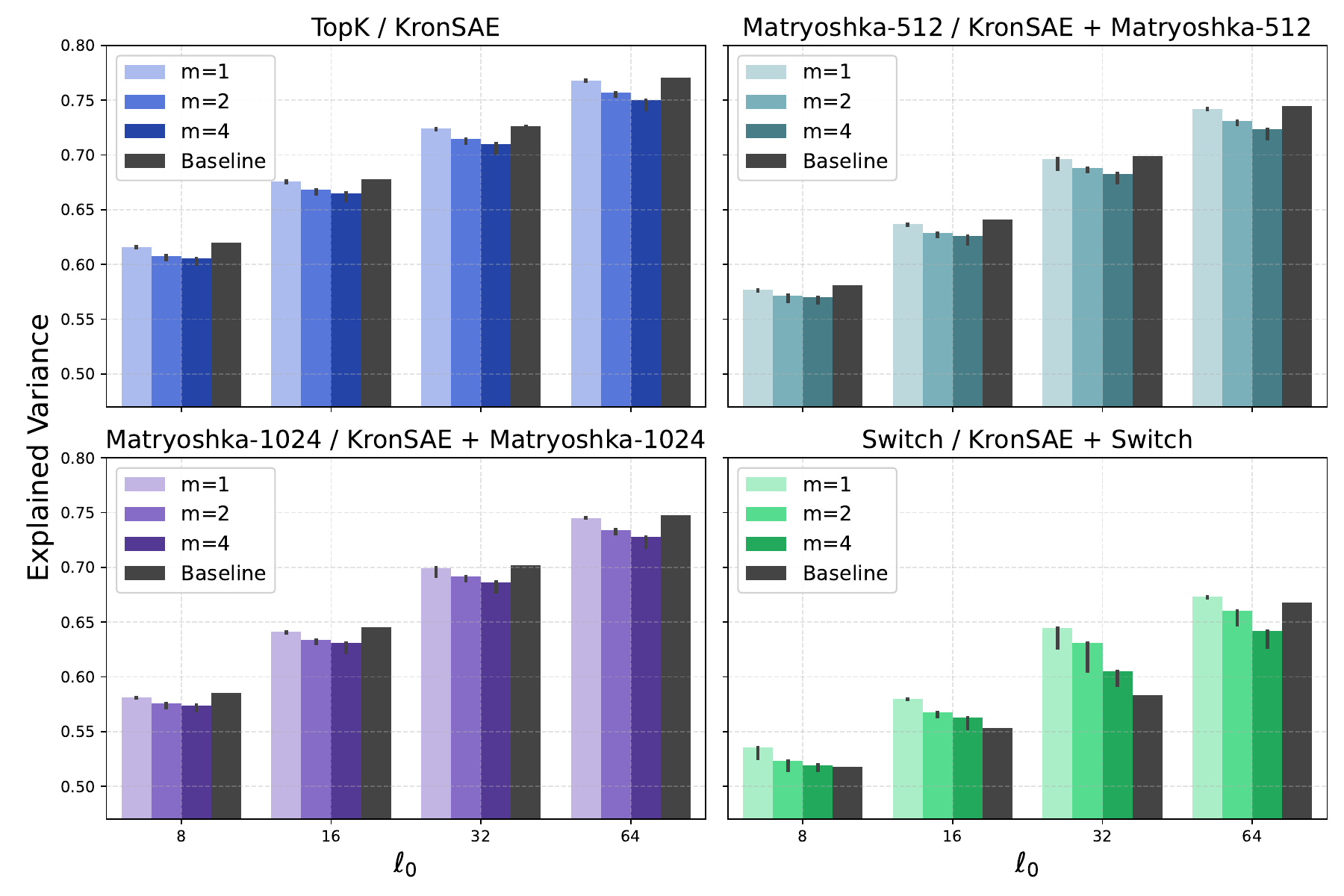}
  \caption{Applying the Kron encoder to TopK, Switch, and Matryoshka SAEs. Switch+Kron matches the Switch baseline in most configurations and improves it in some; Matryoshka+Kron remains on par across the sweep.}
  \label{fig:different_saes}
\end{figure}

\begin{figure}[ht]
    \centering
    \includegraphics[width=1.\linewidth]{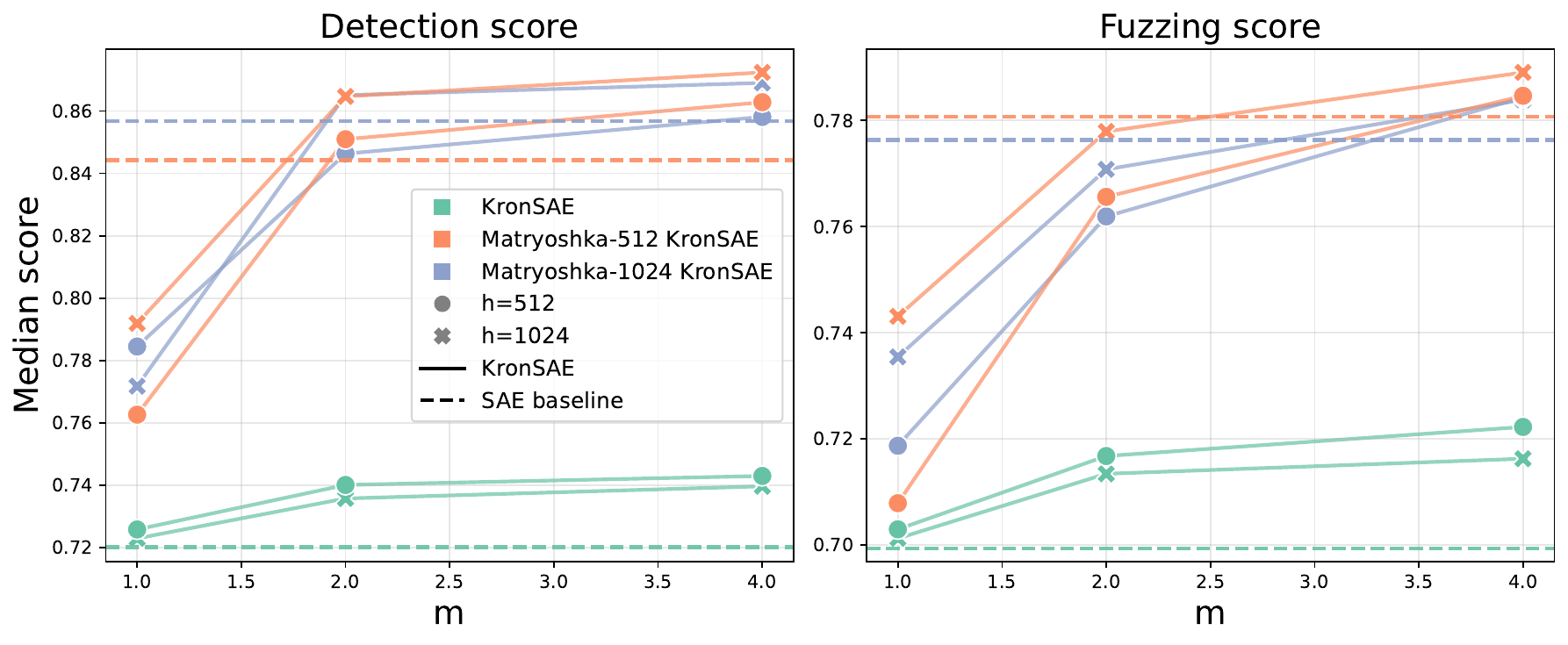}
    \caption{Detection and fuzzing scores for baseline SAEs and their Kron variants. Scores improve with increasing $m$; with $m\ge2$, all Kron variants match or exceed their baselines.}
    \label{fig:interpretability_tradeoff}
\end{figure}

\paragraph{Compatibility with other SAEs.}
The method is compatible with existing SAE variants; we test it with two related architectures on layer 12 of Gemma-2-2B. Combining a Matryoshka SAE \citep{bussmann2025learning} with our encoder improves latent interpretability (Figure~\ref{fig:interpretability_tradeoff}) while maintaining the same representation-quality trade-off (Figure~\ref{fig:different_saes}). For the Switch SAE \citep{mudide2025efficient}, our Kronecker encoder improves quality relative to the baseline (Figure~\ref{fig:different_saes}). At the selected dictionary size, Matryoshka-512 and Matryoshka-1024 have nesting depths of 6 and 5, respectively. Appendix~\ref{sec:details} provides implementation and notation details and reports tests with JumpReLU and Gated SAEs.

\paragraph{Interpretability trade-off.} For this analysis, we select the commonly studied middle layer of both Qwen and Gemma with dictionary size $F=16{,}384$, and use a larger training budget of 1B tokens. We sweep over $m$, $n$, and $h$ to characterize the trade-off among interpretability metrics, explained variance, and KronSAE hyperparameters. As Figure~\ref{fig:interpretability_tradeoff} shows, KronSAE's interpretability scores improve relative to TopK SAE as $m$ increases.

\paragraph{Computational cost.} 

\begin{figure}[ht]
  \includegraphics[width=\columnwidth]{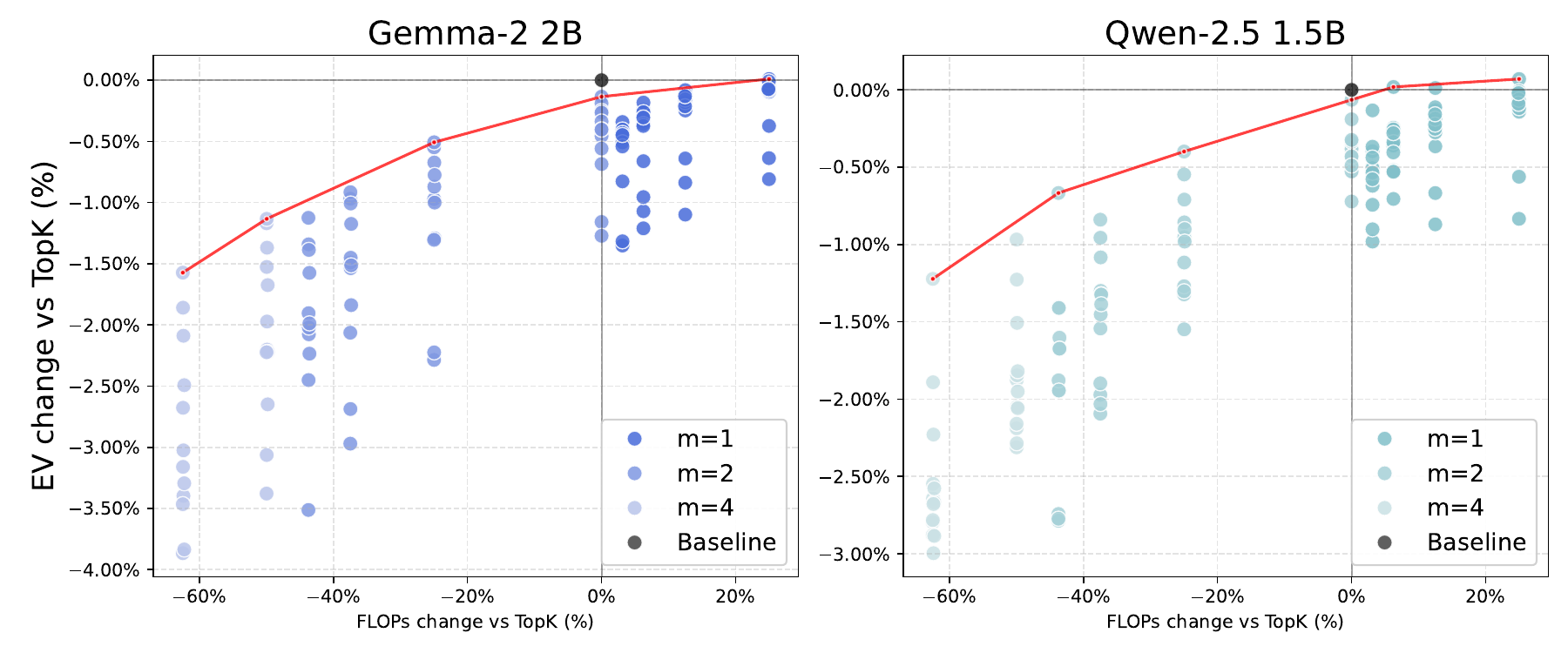}
   \caption{Reconstruction--compute trade-off between TopK SAE and KronSAE. KronSAE reduces encoder FLOPs by more than 40\% while incurring less than a 1\% drop in EV. See Appendix~\ref{sec:computational_cost} and the full comparison in Figure~\ref{fig:speed_comparison_full_plot}.}
  \label{fig:speed_comparison}
\end{figure}

KronSAE also reduces encoder-side computation: it projects the input to only $h(m+n)$ pre-latents instead of directly projecting to all $F=hmn$ post-latents. Figure~\ref{fig:speed_comparison} compares KronSAE configurations from the ablation sweep with the TopK SAE baseline in terms of encoder FLOPs (see Appendix~\ref{sec:computational_cost}) and explained variance. The resulting Pareto curve shows that KronSAE can substantially reduce computation while preserving reconstruction quality.

\section{Study of Correlation Learning} \label{sec:correlation_study}

In this section, we analyze how KronSAE captures correlations between features in a real-world setting and in a controlled toy experiment.

\subsection{Real correlations.}

\begin{figure}[ht]
    \centering
    \includegraphics[width=0.9\linewidth]{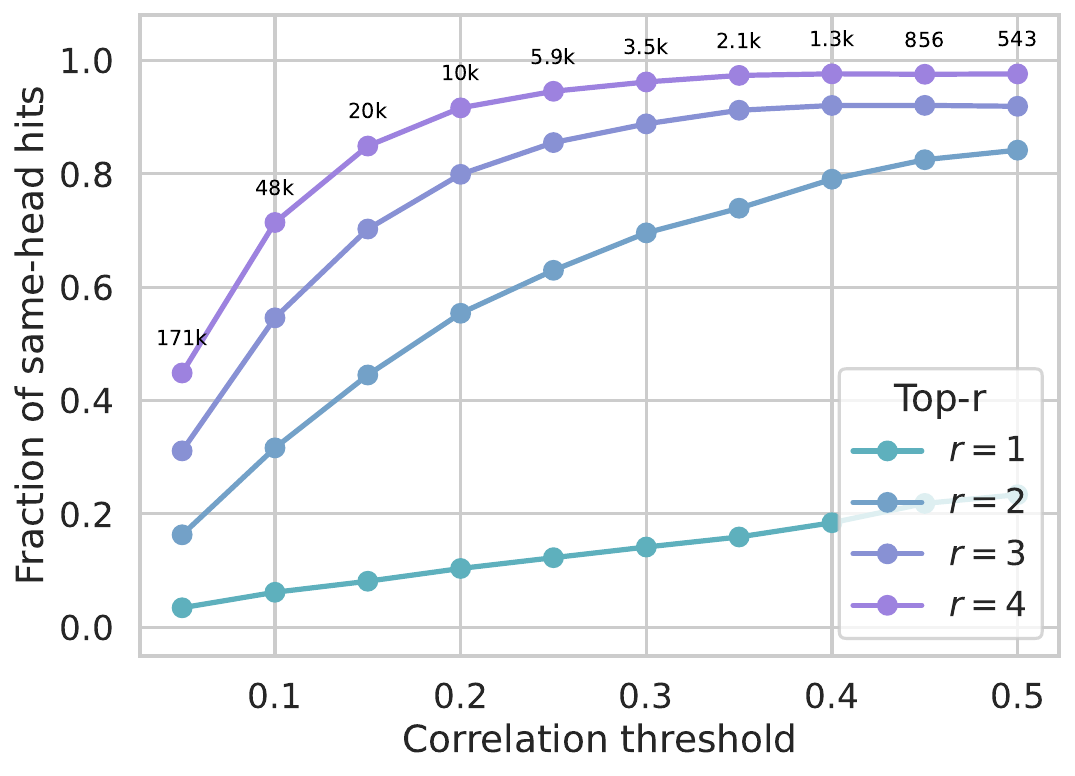}
    \caption{
    Fraction of correlated TopK feature pairs whose top-$r$ KronSAE matches share at least one Kron head.
    We select TopK feature pairs with activation correlation above the threshold on the $x$-axis, map each feature to its top-$r$ correlated KronSAE features, and count a hit if the two candidate sets contain a pair from the same Kron head. All observations are far above the chance value $r^2/h$.
    }
    \label{fig:match_hit_rate}
\end{figure}

To test whether KronSAE heads capture correlated feature structure, we compare a trained KronSAE with a standard TopK SAE trained on the same model activations. We first compute Pearson correlations between TopK feature activations and, separately, between each TopK feature and each KronSAE feature. For every TopK feature $i$, we retain its top-$r$ most correlated KronSAE matches, denoted $\mathcal{M}_r(i)$. We then select TopK feature pairs $(i,j)$ whose activation correlation exceeds a threshold $\tau$, and count a hit when any feature in $\mathcal{M}_r(i)$ and any feature in $\mathcal{M}_r(j)$ belong to the same KronSAE head. The score in Figure~\ref{fig:match_hit_rate} is the fraction of correlated TopK pairs that produce such a same-head hit. This tests whether KronSAE groups correlated features from a flat TopK SAE within the same heads, and whether it does so more often than expected by chance. The chance rate is approximately $1-(1-1/h)^{r^2}\approx r^2/h$, far below the observed rate. Figure~\ref{fig:feature_flow} provides a geometric view of this behavior.

\begin{figure}[ht]
    \centering
    \includegraphics[width=1.0\linewidth]{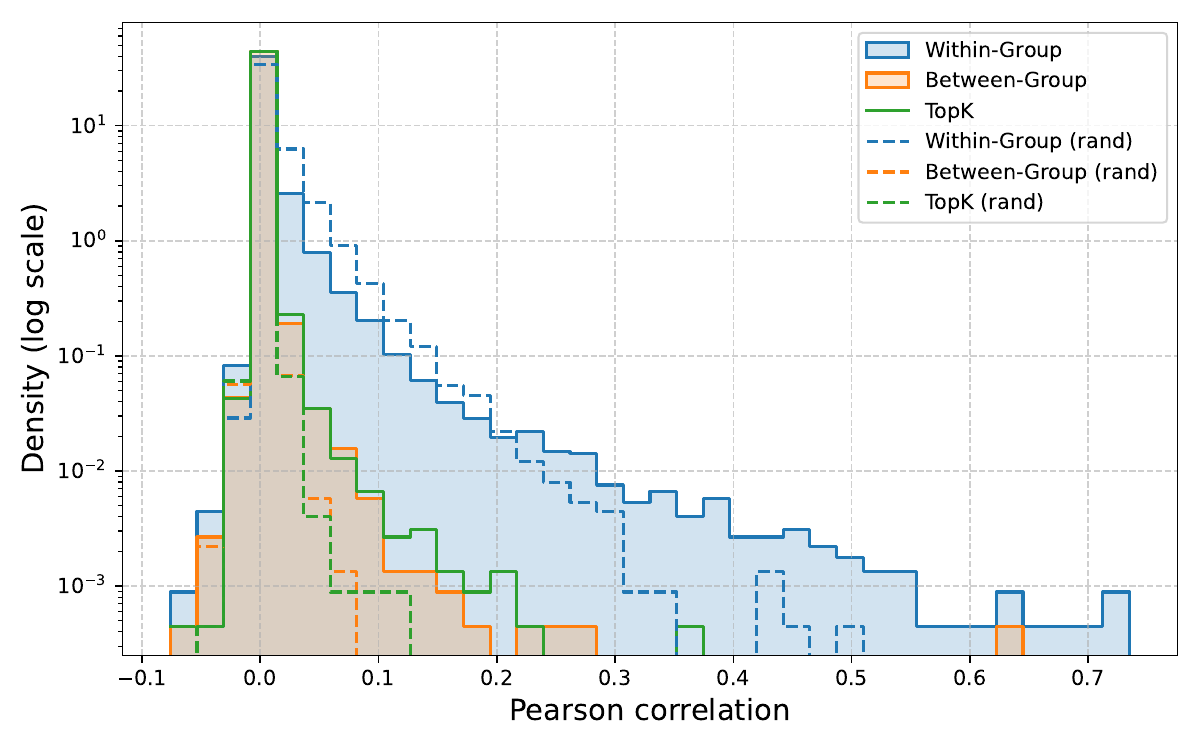}
    \caption{Correlations between features in a KronSAE with $m=4,n=4$, measured within a head and across heads. Our design induces higher within-head correlations, which become stronger after training, although the SAE also learns correlations across heads.}
    \label{fig:correlations}
\end{figure}

To test KronSAE's co-activation inductive bias, we select a KronSAE with $m=4$ and $n=4$ and calculate correlations on 5K texts from the training dataset. For each feature, we compute the mean correlation with features in its head and with all other features, then compare a randomly initialized KronSAE with the trained model. To isolate the effect of our initialization procedure, we sample the weights from a uniform distribution. As Figure~\ref{fig:correlations} shows, correlations are higher within a head and increase further after training, suggesting that the imposed correlation structure works as intended.

\subsection{Toy model of correlations} \label{sec:toy_model_main}

To evaluate how well different sparse autoencoder architectures recover underlying correlation patterns, we construct a controlled sanity-check experiment using a synthetic, block-structured covariance model, where KronSAE's inductive bias is expected to align with the ground-truth structure. We generate sparse nonnegative ground-truth feature activations, linearly mix them using a Cholesky factor $L$ of a block-structured matrix $S = LL^T$, and train a tied-weight bottleneck autoencoder following \citet{elhage2022toymodels}. After training, we match the learned SAE decoder atoms with the AE atoms by solving a quadratic assignment problem \citep{Vogelstein2015FastAQ} (see Appendix~\ref{sec:synthresults} for motivation and details), and compute the decoder-atom similarity matrix $C_{\mathrm{dec}} = W_{\mathrm{dec}}^T W_{\mathrm{dec}}$.

\begin{figure}[tb]
    \centering
    \includegraphics[width=1.0\linewidth]{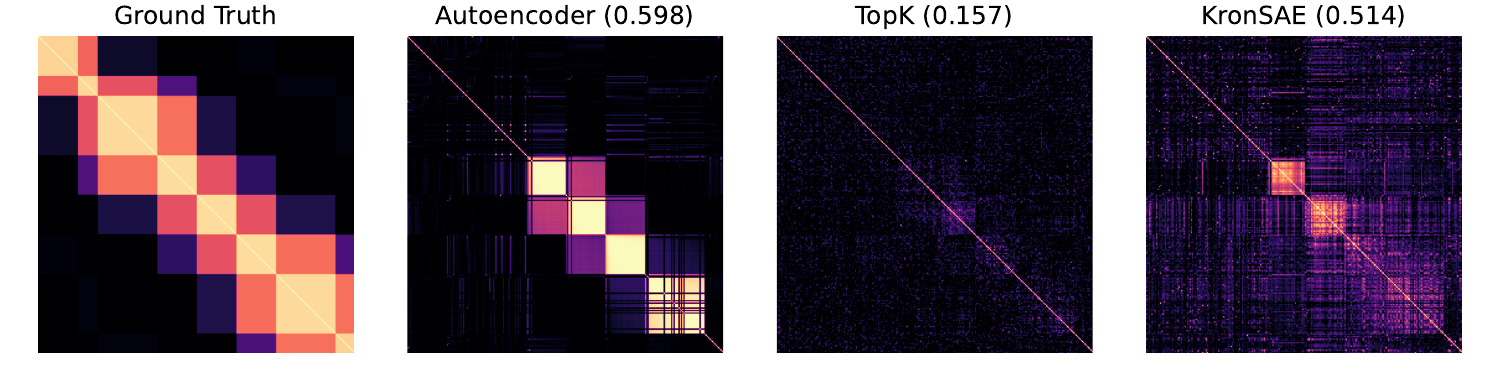}
    \caption{
    Synthetic block-correlation experiment. Each panel shows a feature-similarity matrix for the target block structure, the bottleneck AE representation, TopK SAE, and KronSAE. Numbers in parentheses are RV coefficients against the target after atom matching. Results show that KronSAE recovers the block structure substantially better than TopK SAE. See details in the Appendix \ref{sec:synthresults}.
    }
    \label{fig:synthetic}
\end{figure}

To quantify how closely SAE atoms mirror the original structure of \(S\), we use the RV coefficient, defined as $\operatorname{RV}(A,B) = \operatorname{tr}(A^\top B) / \sqrt{\operatorname{tr}(A^\top A)\operatorname{tr}(B^\top B)}$. Figure~\ref{fig:synthetic} shows the results. KronSAE achieves a higher RV coefficient than TopK SAE in this synthetic setting, confirming that the inductive bias is expressive enough to recover block structure when it is present. Appendix~\ref{sec:synthresults} provides a detailed description and additional setups.

\section{Analysis} \label{sec:experiments}

\subsection{Absorption}
\label{sec:absorption}

A persistent challenge in SAE interpretability is \emph{feature absorption}, where one learned feature becomes a strict subset of another and consequently fails to activate on instances that satisfy the broader concept but not its narrower representation (e.g., a ``starts with L'' feature is subsumed by a ``Lion'' feature) \citep{chanin2024absorption}.

\begin{figure*}[tb]
    \centering
    \includegraphics[width=1.0\linewidth]{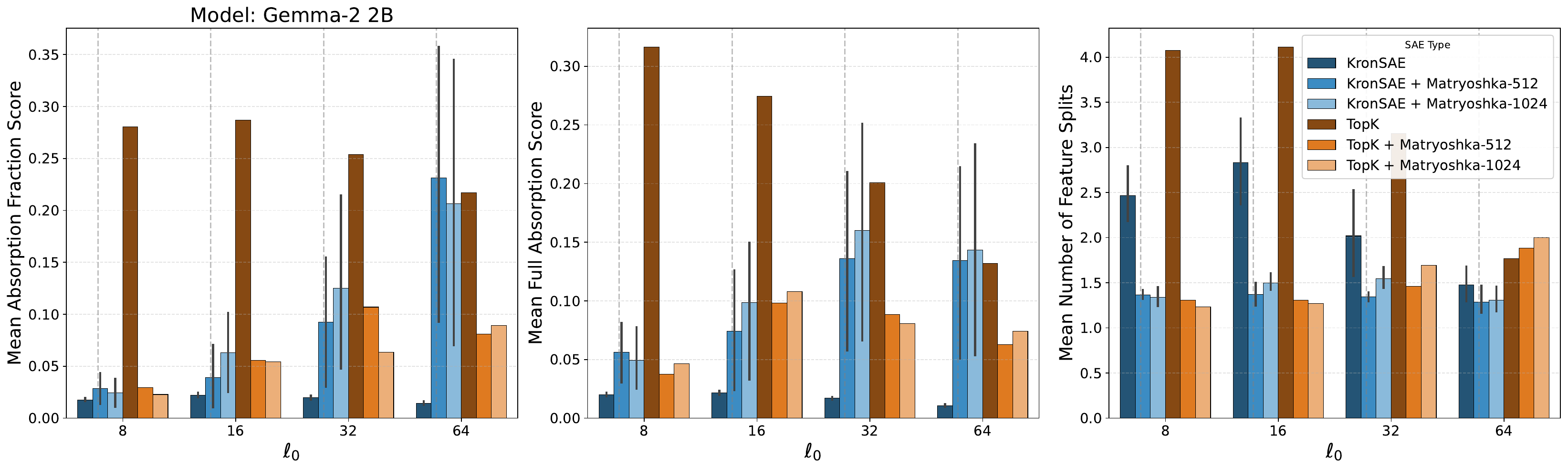}
    \caption{Feature absorption metrics on Gemma-2-2B. KronSAE configurations (various $m,n$) exhibit lower mean absorption fractions and full-absorption scores across different $\ell_0$ values and selected baselines.}
    \label{fig:absorption}
\end{figure*}

Figure~\ref{fig:absorption} reports three absorption metrics measured with SAEBench \citep{karvonen2025saebench} across sparsity levels $\ell_0\in\{8,16,32,64\}$: (1) the \emph{mean absorption fraction}, measuring the proportion of features that are partially absorbed; (2) the \emph{mean full-absorption score}, quantifying complete subsumption events; and (3) the \emph{mean number of feature splits}, indicating how often a single conceptual feature fragments into multiple activations. We use dictionary size $F=2^{14}$ and compare TopK SAE, Matryoshka SAE \citep{bussmann2025learning}, and their KronSAE variants, which impose different hierarchical priors (see Appendix~\ref{sec:logical_operator}). Across all $\ell_0$ values, KronSAE variants tend to reduce the first two scores relative to the TopK SAE baseline while maintaining a similar rate of feature splitting.

We hypothesize that the observed reductions in absorption are consistent with two complementary architectural biases: a mechanism where a post-latent can be active only when both pre-latents are active can reduce full-absorption events by limiting activation when a parent is inactive (see additional description in Appendix \ref{sec:logical_operator}); and dividing the latent space into $h$ independent subspaces (each with its own base and extension factors) encourages correlated features to localize within a head (Section \ref{sec:correlation_study}), which can reduce cross-head absorption.

\subsection{Analysis of Learned Features} \label{sec:featureAnalysis}

In this section, we compare KronSAE and TopK SAE in terms of interpretability and feature properties, and analyze the properties of KronSAE groups. We use layer 12 of Gemma-2-2B and a dictionary size of 16,384 features. The selected KronSAE has $m=4$ and $n=4$. We process 24M tokens to collect the analysis data. Our interpretation pipeline follows a common methodology: an LLM interprets activation patterns \citep{bills2023llmcanexplain}, and we evaluate the resulting interpretations using \emph{detection} and \emph{fuzzing} scores \citep{paulo2024automatically} (see Appendix~\ref{sec:featureAnalysisDetails}).

For each selected feature, in addition to mean activation and frequency, we calculate two metrics. Low \textit{token entropy} suggests that a feature activates frequently on a small number of tokens and is therefore token-specific. A high \textit{multitoken ratio} indicates that a feature tends to activate multiple times in a single sequence. Both metrics have a notable negative correlation with the final interpretability scores and may therefore provide useful proxies without requiring full interpretation-based evaluation.

\begin{figure}[ht]
    \centering
    \includegraphics[width=1\linewidth]{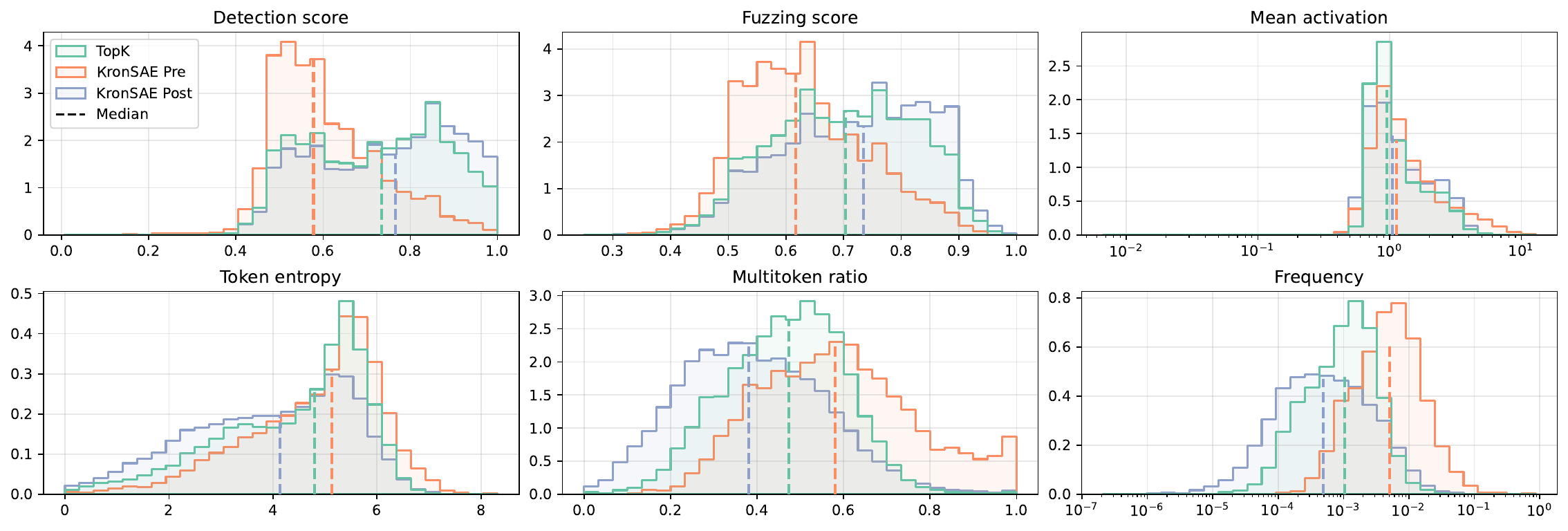}
    \caption{Property distributions for TopK SAE and KronSAE ($m=4,n=4$) with a 16K dictionary trained on Gemma-2-2B. The Pre and Post suffixes denote pre- and post-latents; the $y$-axis shows density. Token entropy is the entropy of the token distribution on which a feature activates, and the multitoken ratio measures how often a feature activates within a single sequence. KronSAE achieves better interpretability scores while learning more specialized feature groups, as indicated by lower activation frequency and lower variance among activated tokens.}
    \label{fig:propertiesDistribution}
\end{figure}

\paragraph{SAE properties and encoding mechanism.} We observe that the features learned by KronSAE are more specific, indicated by lower values of the token entropy and multitoken ratio, and higher interpretability scores than TopK, as shown in Figure \ref{fig:propertiesDistribution}. Since post-latents are significantly more interpretable than corresponding pre-latents, we hypothesize the hidden mechanism for encoding and retrieval of the required semantics.

By examining activating examples and interpretations of latents, we observe that pre-latents may carry multiple distinct and identifiable modes of activation, such as composition \Kbase{base} element 3 in head 23 shown in Table \ref{tab:interactionExample}, and be very abstract compared to resulting post-latents. Polysemanticity of pre-latents is expected to be a consequence of reduced "working" number of encoder latents, since we decompose the full dictionary size and reduce the encoder capacity.

\begin{table*}[ht]
\renewcommand{\arraystretch}{1.2}
\begin{tabularx}{\textwidth}{@{} l X c @{}}
\toprule
Component & Interpretation & Score \\
\midrule
\textbf{\Kbase{Base} 3} & Suffix ``-like'' for comparative descriptors, directional terms indicating geographical regions, and concepts related to spiritual or metaphysical dimensions & 0.84 \\
\midrule
\multicolumn{3}{@{}l}{Extension elements and their compositions with base 3} \\
\midrule
\textbf{\Kext{Extension} 0} & Comparative expressions involving ``than'' and ``as'' in contrastive or proportional relationships. & 0.87 \\
\textit{Composition:} & Similarity or analogy through the suffix ``-like'' across diverse contexts. & 0.89 \\
\addlinespace
\textbf{\Kext{Extension} 1} & Specific terms, names, and abbreviations that are contextually salient and uniquely identifiable. & 0.66 \\
\textit{Composition:} & Medical terminology related to steroids, hormones, and their derivatives. & 0.84 \\
\addlinespace
\textbf{\Kext{Extension} 2} & Spiritual concepts and the conjunction ``as'' in varied syntactic roles. & 0.80 \\
\textit{Composition:} & Abstract concepts tied to spirituality, consciousness, and metaphysical essence. & 0.93 \\
\addlinespace
\textbf{\Kext{Extension} 3} & Directional and regional descriptors indicating geographical locations or cultural contexts. & 0.84 \\
\textit{Composition:} & Directional terms indicating geographical or regional divisions. & 0.91 \\
\bottomrule
\end{tabularx}
\caption{Interactions between composition \Kbase{base} element 3 in head 23 and all \Kext{extension} elements in that head. Interaction happens in a way that closely resembles the Boolean AND operation: \Kbase{base} pre-latent is polysemous, and the composition post-latent is the intersection, i.e. logical AND between pre-latents. See details in Section \ref{sec:featureAnalysis}.}
\centering
\label{tab:interactionExample}
\end{table*}

Thus, we hypothesize that the encoding of specific semantics in our SAE may be done via magnitude, which we validate by examining the activation examples. For the above mentioned pre-latent, the "comparison" part is encoded in the top 75\% quantile, while the "spiritual" part is mostly met in the top 25\% quantile, and the "geographical" part is mainly encoded in the interquartile range. We also consider but do not investigate the possibility that it may depend on the context, e.g. when the model uses the same linear direction to encode different concepts when different texts are passed to it.

\paragraph{Semantic retrieval and interpretable interactions.} \label{par:sem_retrivial_and_interactions}

Heads usually contain groups of semantically related pre-latents, e.g. in head 136 three base elements and one extension covering numbers and ordinality, two extension elements related to geographical and spatial matters, one question-related base and one growth-related extension. 
Retrieval happens primarily through a mechanism resembling a logical AND circuit, where one pre-latent carries multiple semantics and the other acts as a specifier. Table~\ref{tab:interactionExample} gives an illustrative example: the base contains three detectable sub-semantics, and each extension retrieves one of them.

Other types of interaction may occur, including apparently novel semantics (e.g., the composition of base 3 and extension 1 in Table~\ref{tab:interactionExample}). In other cases, a post-latent inherits detectable semantics from only one parent, particularly when the other parent has a broad interpretation and a low score. Identifying the fine-grained structure of these interactions requires more sophisticated techniques, which we leave to future work.

See more examples in Appendix \ref{sec:compositionalStructure}. For details on dependencies between feature properties and feature geometry, see Appendix \ref{sec:additionalAnalysis}.

\section{Conclusion}
\label{sec:conclusion}

We introduce \textbf{KronSAE}, a sparse autoencoder architecture that combines head-wise Kronecker factorization of the latent space with an approximation of logical AND through the $\operatorname{mAND}$ nonlinearity, inducing a co-activation prior on latent structure. Our approach enables efficient training of compositional SAEs, especially with limited compute or training data, while maintaining reconstruction fidelity and yielding more interpretable features. Our analysis links these gains to the complementary effects of compositional latent structure and logical AND-style interactions, offering a new perspective on how sparsity and factorization can work together in representation learning.

\section*{Limitations}
KronSAE introduces a clear trade-off among interpretability, efficiency, and reconstruction performance. Our evaluation is limited to mid-sized transformer models. Although the overall training and inference infrastructure is similar to that of standard SAEs, replacing the dense encoder with the proposed factorized encoder requires additional implementation work. The resulting computation is highly parallelizable and is already reasonably efficient with standard tensor operations, even without specialized kernels. We therefore view this engineering overhead as moderate relative to KronSAE's interpretability and compute trade-offs. Finally, the toy-model setup is intended primarily as a sanity check of KronSAE's inductive bias; real-world correlation and hierarchical structure may differ. Our real-data correlation experiment provides a more direct, practically oriented evaluation.

\bibliography{custom}

\newpage
\appendix

\section{Experimental Setup Details} \label{sec:details} 

\paragraph{Training details.}
All SAEs are optimized using AdamW with an initial learning rate of $8\times10^{-4}$, a cosine learning-rate schedule with a minimum learning rate of $1\times10^{-6}$, a linear warmup over the first 10\% of training steps, and an auxiliary-loss coefficient of $0.03125$. We use a global batch size of 8,192. We sweep over dictionary sizes $F\in\{2^{14},2^{15},2^{16}\}$. For KronSAE, we also sweep the number of heads $h$ and per-head dimensions $m$ and $n$, subject to $hmn=F$. Regularization weights and auxiliary-loss coefficients are held constant across runs to isolate the effect of architecture.

Across all experiments, including preliminary research, we used approximately 330 GPU-days on NVIDIA H100 80GB GPUs.

\paragraph{Initialization.}
As observed by \citet{gao2025scaling}, initializing the decoder as the transpose of the encoder ($W_{\mathrm{dec}}=W_{\mathrm{enc}}^\top$) substantially improves performance. We adapt this strategy to KronSAE by partitioning $W_{\mathrm{enc}}$ into $h$ head-wise pairs of blocks with shapes $m\times d$ and $n\times d$, denoted $\{\Kbase{P_k},\Kext{Q_k}\}_{k=1}^h$. For each head $k$, we define decoder rows through additive composition:
\[
C_k[i,j] \;=\; \Kbase{P_{k,i}} \;+\; \Kext{Q_{k,j}},
\]
with $i=1,\dots,m$ and $j=1,\dots,n$. Flattening the matrices $\{C_k\}$ yields the full decoder weight matrix $W_{\mathrm{dec}}\in\mathbb{R}^{F\times d}$ in the row-vector convention used by our implementation.

\paragraph{Matryoshka and Kron-based version.} For a Matryoshka SAE \citep{bussmann2025learning} with dictionary size $F$, we use the training settings above and define dictionary groups $S=[2^k,2^k,\ldots,2^{k+i},\ldots,2^n]$, where $k<n$ and $\sum_{s\in S}s=F$. This is equivalent to nested sub-SAEs with dictionary sizes $\mathcal{M}=\{2^k,2^k+2^k,2^k+2^k+2^{k+1},\ldots,F\}$. We refer to this model as \textbf{Matryoshka-}$\mathbf{2^k}$. Its reconstruction loss is
\begin{equation}
\mathcal{L}(\mathbf{f}) = \sum_{m \in \mathcal{M}} \left\lVert\mathbf{x}-\left(\mathbf{f}_{0:m}W_{0:m}^{\mathrm{dec}}+\mathbf{b}_{\mathrm{dec}}\right)\right\rVert^2_2,
\end{equation}
where $\mathbf{f}$ is the latent vector. During training, we also use the auxiliary loss $\mathcal{L}_{\mathrm{aux}}$ from \citet{gao2025scaling}.

For the Kron-based Matryoshka SAE, we use the same setup and initialization. We choose $m$ and $n$ so that $F \bmod (mn)=0$, ensuring that no heads are shared between groups.

\paragraph{Switch SAE and Kron-based version.} For the Switch SAE architecture proposed by \citet{mudide2025efficient}, we use 8 experts in all experiments and the same initialization as KronSAE. In the Kron-based variant, we distribute the heads equally across experts, yielding $F/(8mn)$ heads per expert. This reduces both the number of parameters in each expert and encoder FLOPs.

\paragraph{Comparison with JumpReLU and Gated SAE.}
We compare KronSAE with alternative activation mechanisms, including JumpReLU and Gated SAE, and report explained variance at three sparsity levels in Table~\ref{tab:gated_jumprelu_sae}. For Gated SAEs, the reconstruction performance of Gated-KronSAE is on par with that of the baseline.

\section{Computational Cost} \label{sec:computational_cost}

\begin{figure}[ht]
    \centering
    \includegraphics[width=1\linewidth]{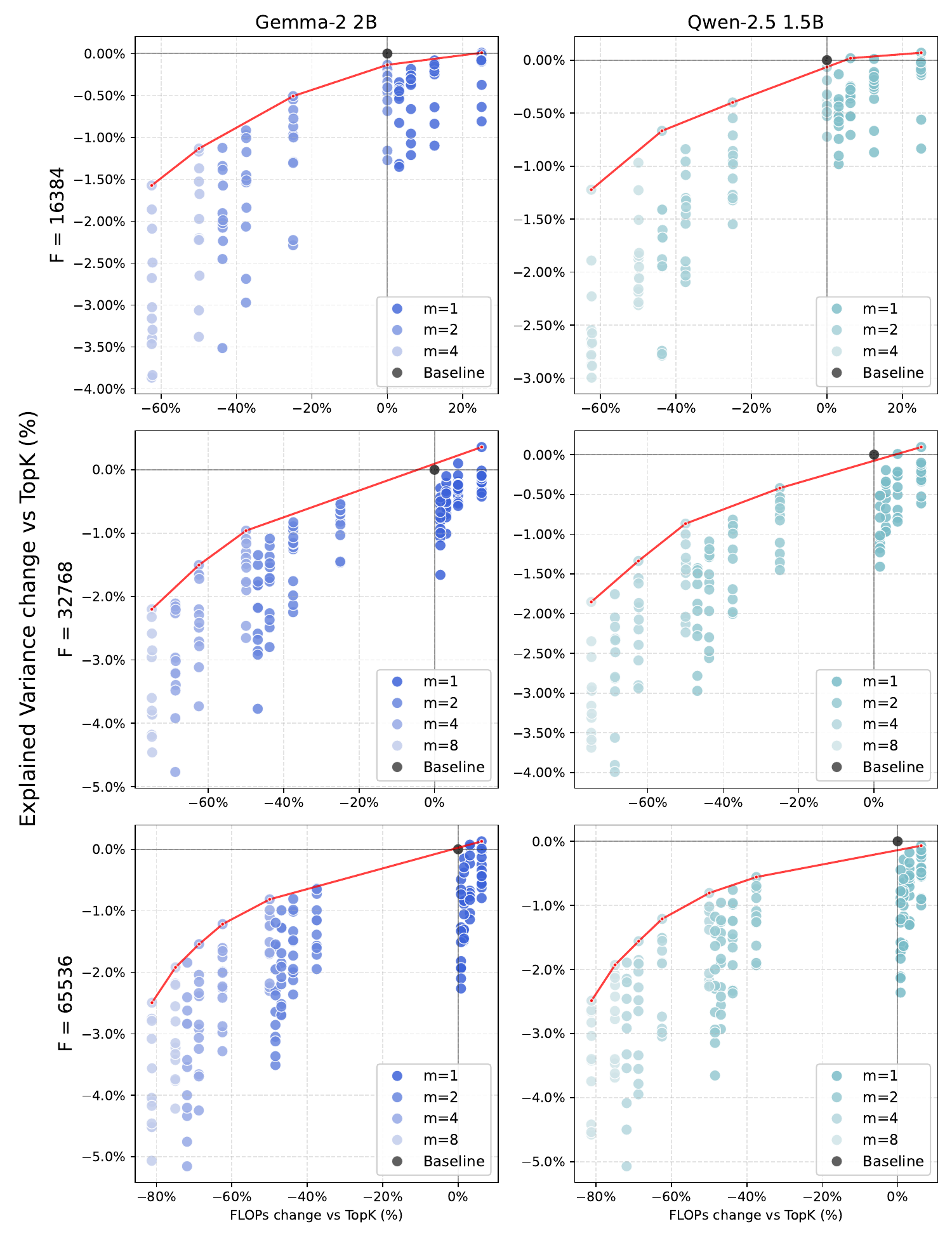}
    \caption{Trade-off between FLOPs and explained variance. In most settings, KronSAE substantially reduces compute with only a small change in EV.}
    \label{fig:speed_comparison_full_plot}
\end{figure}

\begin{table}[t]
    \centering
    \small
    \setlength{\tabcolsep}{8pt}
    \begin{tabular}{c|c|c|c}
        \toprule
        SAE & $\ell_0=32$ & $\ell_0=50$ & $\ell_0=64$ \\
        \toprule
        TopK & 0.809 & 0.837 & 0.852 \\
        JumpReLU & 0.813 & 0.838 & 0.844 \\
        Kron+TopK & 0.814 & 0.840 & 0.853 \\
        Kron+JumpReLU & 0.790 & 0.817 & 0.828 \\
        \bottomrule
    \end{tabular}

    \vspace{0.8em}

    \small
    \setlength{\tabcolsep}{8pt}
    \begin{tabular}{l c c}
        \toprule
        SAE & {$\ell_0$} & {EV $\uparrow$} \\
        \midrule
        Kron+Gated ($m{=}1,n{=}32$) & 63 & 0.8319 \\
        Gated (run 1)            & 68 & 0.83050 \\
        Kron+Gated ($m{=}1,n{=}16$) & 63 & 0.83027 \\
        Kron+Gated ($m{=}2,n{=}8$)  & 66 & 0.82060 \\
        Gated (run 2)            & 63 & 0.81300 \\
        \bottomrule
    \end{tabular}
    \caption{Activation-function ablation for KronSAE. \textbf{Top}: performance of SAEs with JumpReLU and TopK activations. Because JumpReLU has variable sparsity controlled by an $\ell_0$ penalty coefficient, we sweep over sparsity levels, fit a quadratic curve to the resulting EV values as a function of $\ell_0$, and evaluate the curve at the reported levels. TopK and KronSAE with TopK are instead trained at fixed, predefined sparsity levels. \textbf{Bottom}: comparison of the Gated SAE architecture with its KronSAE variant for dictionary size $F=2^{16}$.}
    \label{tab:gated_jumprelu_sae}
\end{table}

We estimate the per-token floating-point operation (FLOP) cost of the SAE forward pass under the standard TopK decoding implementation: the encoder is a dense projection, while the decoder reconstructs using only the $k$ active latents.

For a hidden dimension $d$, dictionary size $F$, and sparsity level $k$, the TopK SAE cost is
\begin{equation}
\mathrm{FLOPs}_{\mathrm{TopK}} = dF + kd = d(F+k),
\end{equation}
where $dF$ corresponds to encoder matrix multiplication and $kd$ corresponds to sparse decoding using the $k$ selected features.

For KronSAE with $h$ heads and factor sizes $(m,n)$ (so $F = hmn$ post-latents), the encoder computes $(m+n)$ pre-latents per head via dense projections and then forms $mn$ post-latents per head via the mAND interaction. This yields
\begin{equation}
\begin{gathered}
\mathrm{FLOPs}_{\mathrm{Kron}}
= dh(m+n) + mnh + kd \approx \\
\approx dh(m+n) + kd = d\bigl(h(m+n)+k\bigr),
\label{equation:flops_calc_formula}
\end{gathered}
\end{equation}
where the $mnh=F$ term accounts for element-wise interactions over all $(i,j)$ pairs in each head. In our typical settings, $d\,h(m+n)$ dominates $F$, so the approximation is accurate up to a small additive $\mathcal{O}(F)$ term.

\begin{figure}[ht]
    \centering
    \includegraphics[width=1\linewidth]{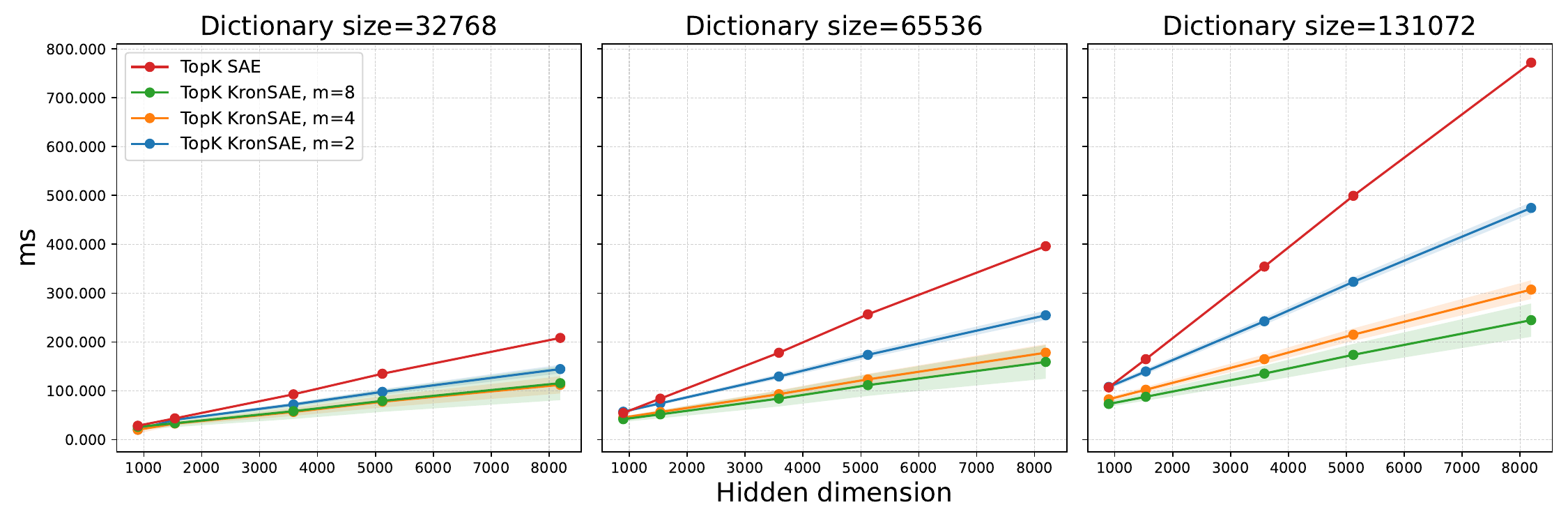}
    \caption{Training-step time for TopK SAE and KronSAE across hidden dimensions. KronSAE scales more favorably than an SAE with the standard encoder architecture.}
    \label{fig:wallclockScaling}
\end{figure}

To corroborate the analytic scaling, we additionally benchmark the wall-clock time of a full SAE training step (forward + backward) for both methods under identical hardware and batching. Figure~\ref{fig:wallclockScaling} reports how step time scales with the hidden dimension $d$.

\section{Performance on Different Layers}\label{appendix:different_layers_ablation}
\begin{figure}[ht]
    \centering
    \includegraphics[width=\linewidth]{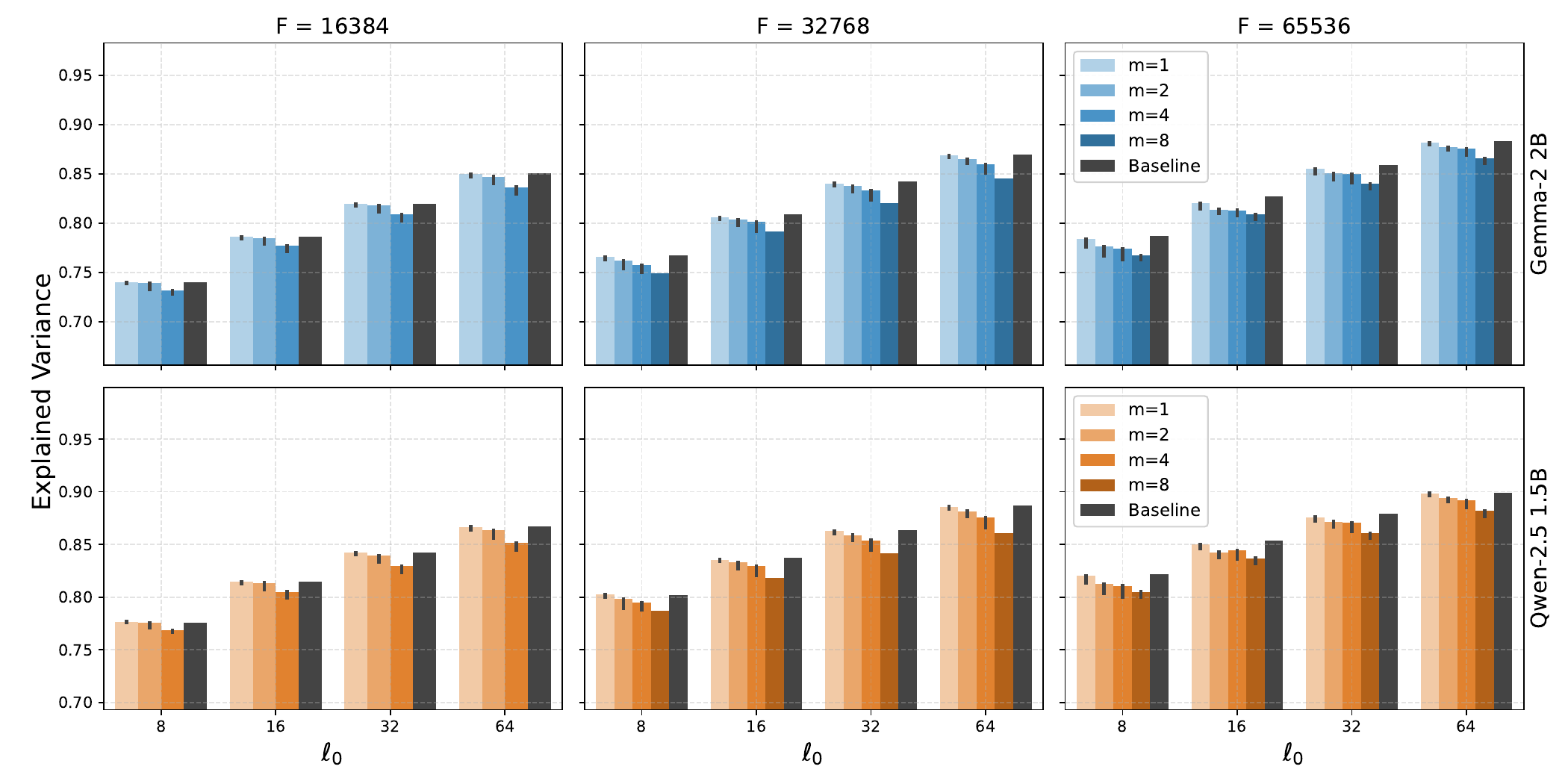}
    \caption{Maximum performance for KronSAE and TopK SAE on Qwen2.5-1.5B and Gemma-2-2B at different dictionary sizes $F$ under an equal token budget. KronSAE uses fewer encoder parameters while remaining on par with the baseline in EV; the gap narrows at larger dictionary sizes on \textbf{layer 6} of both models.}
    \label{fig:ablation_l6_maximum_performance}
\end{figure}

\begin{figure}[ht]
    \centering
    \includegraphics[width=\linewidth]{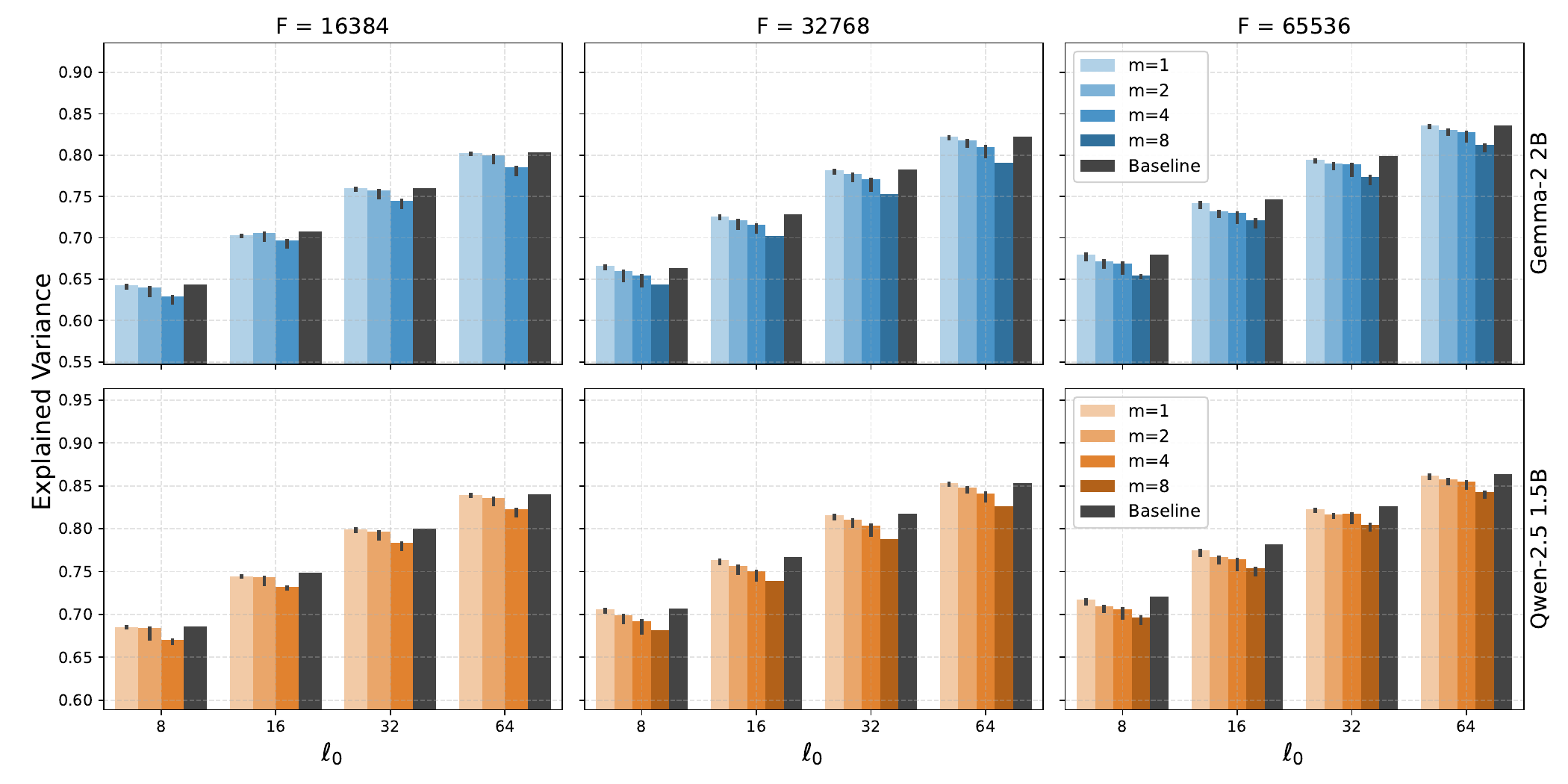}
    \caption{Maximum performance for KronSAE and TopK SAE on Qwen2.5-1.5B and Gemma-2-2B at different dictionary sizes $F$. KronSAE uses fewer encoder parameters while remaining on par with the baseline in EV; the gap narrows at larger dictionary sizes on \textbf{layer 20} of both models.}
    \label{fig:ablation_l20_maximum_performance}
\end{figure}

Figures~\ref{fig:ablation_l6_maximum_performance} and~\ref{fig:ablation_l20_maximum_performance} show that KronSAE is on par with TopK SAE when $m=1$. Other configurations incur only small EV changes while substantially reducing the number of encoder parameters.

\subsection{MSE Performance}\label{sec:mse_performance}
\begin{figure*}[ht]
    \centering
    \includegraphics[width=\linewidth]{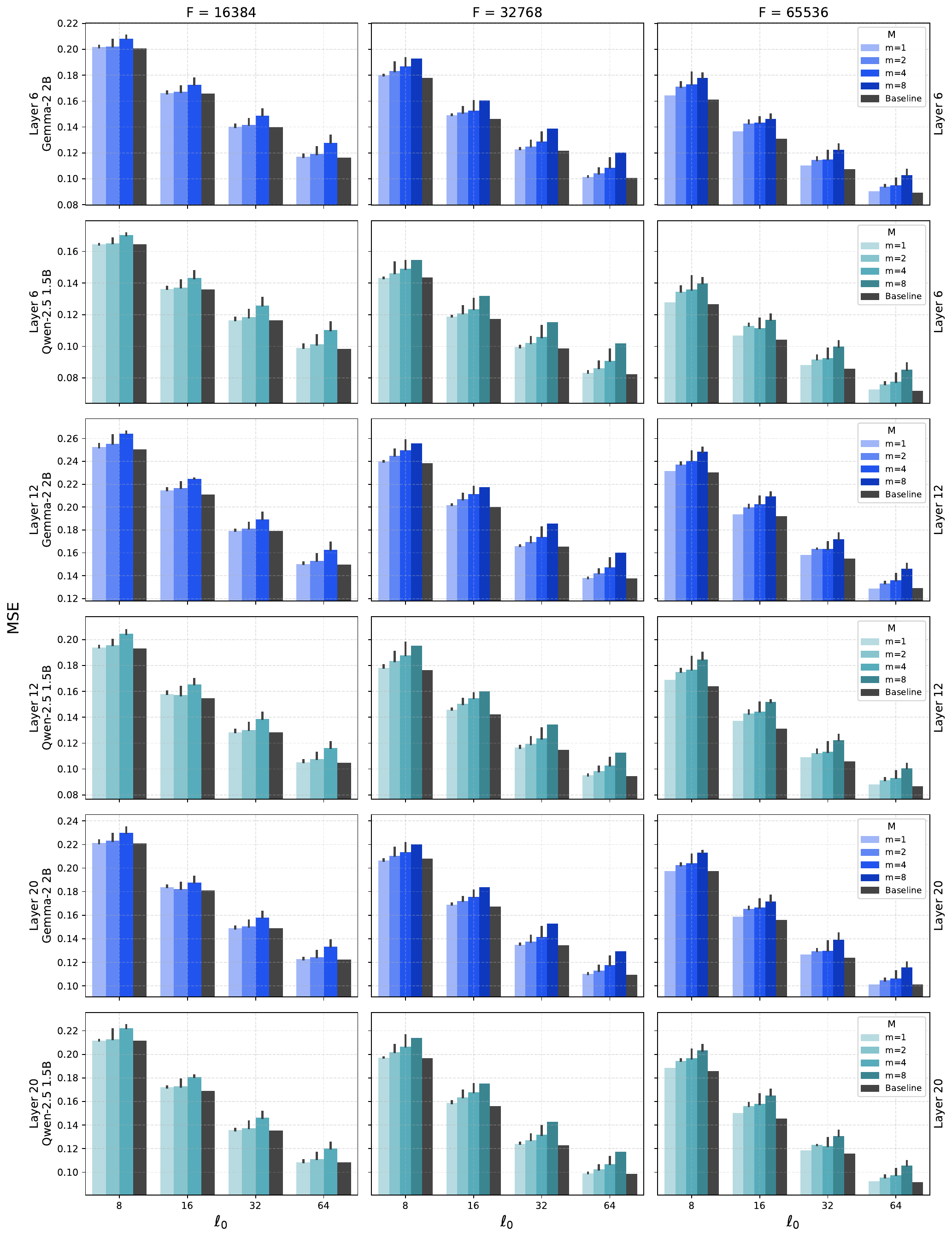}
    \caption{MSE-based comparison of KronSAE and TopK SAE under a fixed token budget.}
    \label{fig:mse_performance_all_layers}
\end{figure*}

We also report MSE results for layers 6, 12, and 20 in Figure~\ref{fig:mse_performance_all_layers}. The trends are consistent with the explained-variance results: under the same fixed token budget, KronSAE matches the reconstruction quality of the TopK SAE baseline across the evaluated sparsity regime.

\section{mAND as a Logical Operator and KronSAE as a Logical SAE}
\label{sec:logical_operator}

Suppose we have $\Kbase{\mathbf{u}^k} = \Kbase{P^k}\mathbf{x}$ and $\Kext{\mathbf{v}^k} = \Kext{Q^k} \mathbf{x}$. KronSAE can be described in two equivalent ways:
\begin{enumerate}
    \item Apply $\operatorname{ReLU}$ to latents $\Kbase{\mathbf{u}}$ and $\Kext{\mathbf{v}}$, then apply the Kronecker product and square root.
    \item Apply the $\operatorname{mAND}$ kernel to create the matrix in Equation~\ref{eq:mand}, then flatten it.
\end{enumerate}
These descriptions are equivalent. To understand how they induce the AND-like mechanism, consider base and extension pre-latent vectors $\Kbase{\mathbf{p}} = \operatorname{ReLU}(\Kbase{\mathbf{u}})$ and $\Kext{\mathbf{q}} = \operatorname{ReLU}(\Kext{\mathbf{v}})$. Here, $\Kbase{p_i}$ is the $i$th base activation and $\Kext{q_j}$ is the $j$th extension activation. The Kronecker product $\Kbase{\mathbf{p}} \otimes \Kext{\mathbf{q}}$ creates the post-latent activation vector $(\Kbase{p_1}\Kext{q_1},\ldots,\Kbase{p_1}\Kext{q_n},\Kbase{p_2}\Kext{q_1},\ldots,\Kbase{p_m}\Kext{q_n})$. Therefore:
\begin{enumerate}
    \item A post-latent is active $\implies$ both parent pre-latent activations are positive.
    \item A post-latent is inactive $\implies$ at least one parent pre-latent activation is zero.
\end{enumerate}

Hence, a post-latent is active only when both parent pre-latents are active. Fix a post-latent and its parents, and let $\Kbase{\mathcal{P}}$, $\Kext{\mathcal{Q}}$, and $\mathcal{F}$ be the sets of hidden-state inputs on which the base pre-latent, extension pre-latent, and post-latent are active, respectively. Then $\mathcal{F}\subseteq\Kbase{\mathcal{P}}\cap\Kext{\mathcal{Q}}$. Thus, pre-latents can be broader and more polysemous than their emitted post-latents. We validate this behavior qualitatively in Section~\ref{sec:featureAnalysis}, where we also observe interactions that may exploit activation magnitude and do not appear to be simple intersections.

\paragraph{Matryoshka and Kron hierarchies.} In contrast to TopK SAE, the Matryoshka loss imposes a hierarchy by dividing the dictionary into groups of $G_1,\ldots,G_k$ latents at different levels of granularity. The first $G_1$ latents are broad and abstract; each subsequent group adds finer-grained semantics. The loss requires progressively larger prefixes to reduce reconstruction error while the smallest prefix remains useful, but it does not require conditional feature activation. KronSAE instead imposes a two-level AND-like hierarchy through an architectural co-activation prior. Because these mechanisms act through the loss and encoder architecture, respectively, they can be combined, as shown in Sections~\ref{sec:ablations} and~\ref{sec:absorption}.

\begin{figure}[ht]
    \centering
    \includegraphics[width=.85\linewidth]{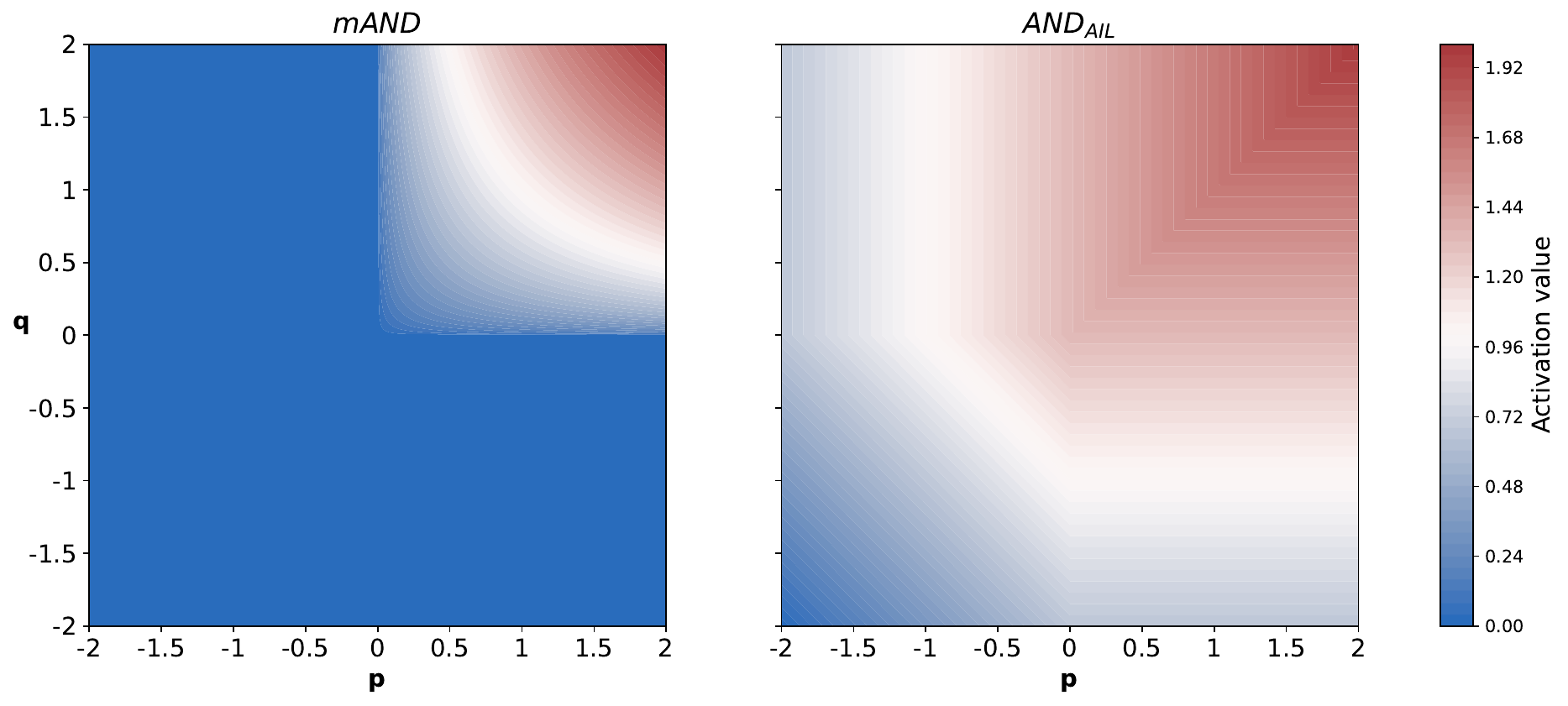}
    \caption{Comparison of the smooth $\operatorname{mAND}$ operator with $\operatorname{AND_{AIL}}$~\citep{Lowe2021}.}
    \label{fig:mand_vis}
\end{figure}

\paragraph{Visual intuition.} We also compare $\operatorname{mAND}$ with $\operatorname{AND_{AIL}}$~\citep{Lowe2021}. Because our objective is to drive each atom toward a distinct, monosemantic feature, tightening the logical conjunction can encourage sharper feature separation. Moreover, using the geometric mean ($\sqrt{\Kbase{p}\,\Kext{q}}$) instead of a simple product or minimum moderates the scale of the post-latent activation when both $\Kbase{p}$ and $\Kext{q}$ are positive. Figure~\ref{fig:mand_vis} provides a visual comparison.

\section{Synthetic Correlations Experiment} \label{sec:synthresults}

This section describes the setup and metrics for the toy-model study of correlation recovery.

\subsection{Correlated Toy Model}

Input vectors $\mathbf{x} \in\mathbb{R}^{F}$ are sampled from a normal distribution (with $\mu=0, \sigma=1$). We then perform a Cholesky decomposition $S = LL^{\top}$ on the covariance matrix \(S\) and set $\bar{\mathbf{x}}_{\mathrm{sparse}} = L\, \operatorname{TopK}(\operatorname{ReLU}(\mathbf{x}))$, so that $\bar{\mathbf{x}}_{\mathrm{sparse}}$ exhibits the desired structure.

Following \citet{elhage2022toymodels}, we train an autoencoder (AE) to reconstruct $\bar{\mathbf{x}}_{\mathrm{sparse}}$:
\begin{equation}
    \mathbf{\hat x} = \operatorname{ReLU}(W^\top W \cdot \bar{\mathbf{x}}_{\mathrm{sparse}} + \mathbf{b}).
     \label{equation:toymodel}
\end{equation}

We collect $d=64$ dimensional hidden states ($W\bar{\mathbf{x}}_{\mathrm{sparse}}$) from the AE, then train a TopK SAE and KronSAE with $F=256$ and $K=8$ to reconstruct them. After training, we extract each SAE's decoder weight matrix $W_{\mathrm{dec}}$, match its latents to the AE latents by solving the quadratic assignment problem \citep{Vogelstein2015FastAQ} described below, and compute the decoder-atom similarity matrix $C_{\mathrm{dec}}=W_{\mathrm{dec}}W_{\mathrm{dec}}^\top$.

\subsection{Quadratic Matching Algorithm}

We next motivate our matching algorithm and present additional results.

Suppose two sets of feature embeddings from different models are represented by matrices $X,Y\in\mathbb{R}^{F\times d}$, where $F$ is the number of features and $d$ is the embedding dimension. Our task is to find an assignment from features in $Y$ to features in $X$ that best preserves their global structure.

The standard approach solves a linear assignment problem: find a permutation matrix $\Pi$ that minimizes $\operatorname{tr}(C^\top\Pi)$, where $C_{i,j}=d(X_i,Y_j)$ is a pairwise feature-distance matrix. The Hungarian algorithm is commonly used for this purpose in sparse dictionary learning \citep{paulo2025sparseautoencoderstraineddata,balagansky2025mechanistic,fel2025archetypal}.

This linear problem considers pairwise feature information but ignores dependencies within the $X$ and $Y$ sets; for example, clusters should map to clusters. We therefore seek a permutation $\Pi$ that maximizes \emph{global} alignment, measured by the squared Frobenius norm of $X^\top\Pi Y$:
\begin{equation} \label{eq:quadraticMatching}
\begin{gathered}
    \max_{\Pi}\|X^\top \Pi Y\|^2_{F} = \\ = \max_{\Pi}\operatorname{tr}(\Pi^\top X X^\top \Pi Y Y^\top),
\end{gathered}
\end{equation}
where global feature structure is explicitly encoded in the matrices $X X^T$ and $Y Y^T$.

We solve this quadratic assignment problem using the Fast Approximate Quadratic assignment (FAQ) method \citep{Vogelstein2015FastAQ}, originally developed for graph matching. Given adjacency matrices $A$ and $B$, FAQ relaxes $\Pi$ from the set of permutation matrices to the set of doubly stochastic matrices (the Birkhoff polytope) and optimizes the corresponding trace objective. In our case, $A=XX^\top$ and $B=YY^\top$, and we maximize similarity:
\begin{equation}
\begin{gathered}
\max_{\Pi}\operatorname{tr}(XX^\top \Pi YY^\top \Pi^\top) = \\ = \max_{\Pi}\operatorname{tr}(\Pi^\top XX^\top \Pi YY^\top),
\end{gathered}
\end{equation}
which is equivalent to Equation~\ref{eq:quadraticMatching}. This formulation preserves global dependencies because the contribution of assigning $Y_j$ to $X_i$ depends on all other assignments through the cross-terms in the quadratic form. Listing~\ref{listing:assignment} shows the matching procedure.

\begin{lstlisting}[
    style=aclpython,
    caption={Implementation of FAQ algorithm for quadratic feature assignment problem.},
    label={listing:assignment}
]
def feature_matching(A, B, max_iter):
    F, d = A.shape
    G_A, G_B = A @ A.T, B @ B.T
    P = np.ones((F, F)) / F # Barycenter initialization
    
    # Frank-Wolfe iterations
    for _ in range(max_iter):
        grad = 2 * G_A @ P @ G_B  # Gradient for maximization
        r, c = linear_sum_assignment(-grad)  # Solve LAP
        Q = np.zeros_like(P)
        Q[r, c] = 1
        
        # Compute optimal step size
        D = Q - P
        a = np.trace(G_A @ D @ G_B @ D.T)
        b = np.trace(grad.T @ D)
        
        if abs(a) < 1e-12:
            alpha = 1.0 if b > 0 else 0.0
        elif a < 0:  # Concave case
            alpha = np.clip(-b/(2*a), 0, 1)
        else:        # Convex case
            alpha = 1.0 if b > 0 else 0.0
            
        P_new = P + alpha * D
        P = P_new
    
    # Project to permutation matrix
    r, c = linear_sum_assignment(-P)
    P_final = np.zeros((F, F))
    P_final[r, c] = 1
    
    return P_final, P_final @ B
\end{lstlisting}

\subsection{Results}

\begin{figure*}[ht]
    \centering
    \includegraphics[width=1\linewidth]{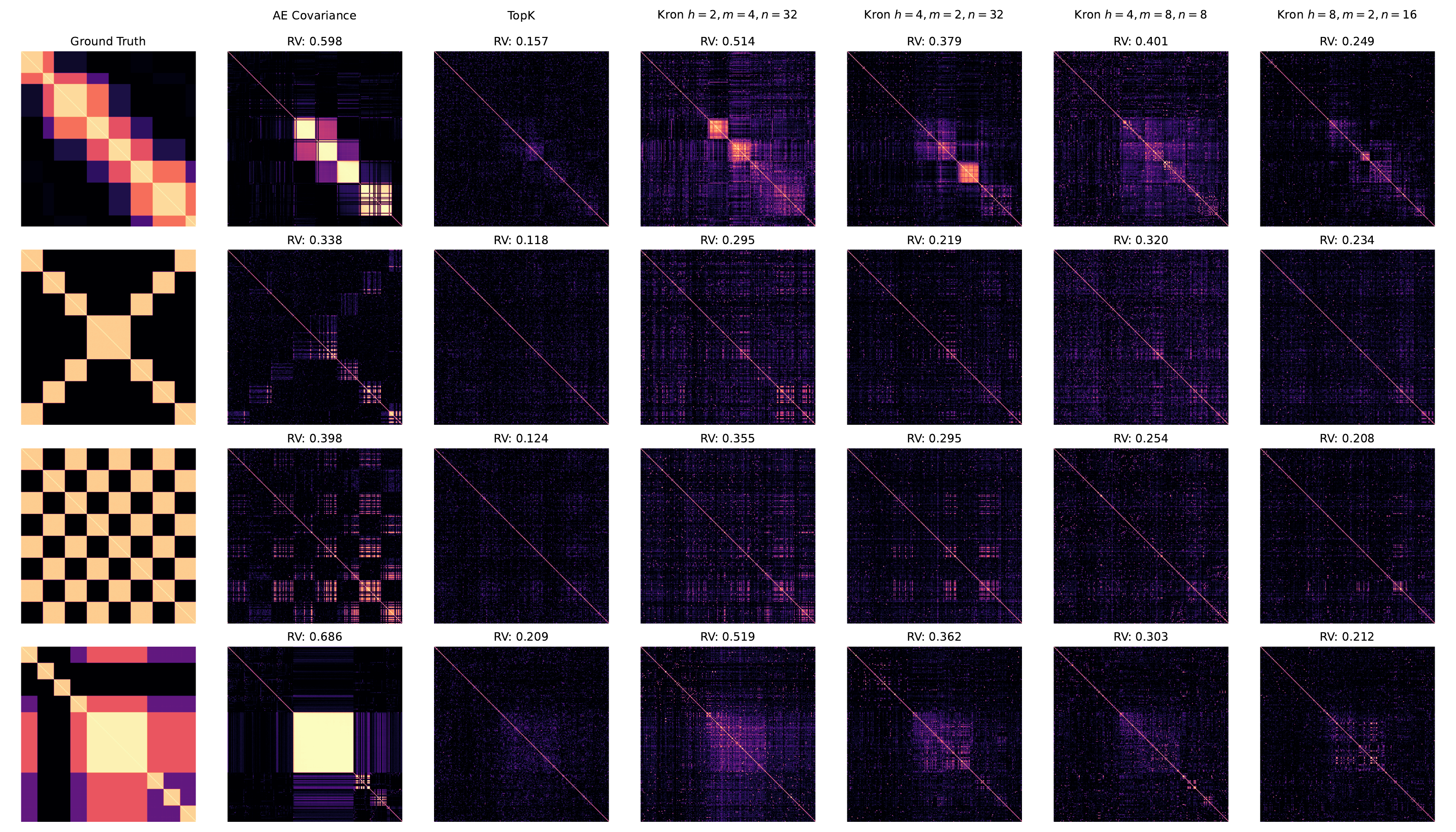}
    \caption{Patterns learned by the autoencoder, TopK SAE, and KronSAE variants after quadratic matching. KronSAE learns patterns that more closely resemble the ground-truth structure; as the number of heads increases and groups become more fine-grained, its behavior approaches that of TopK SAE.}
    \label{fig:patternsImproved}
\end{figure*}

To quantify how closely each model's feature correlations mirror the original structure of \(S\), we use the RV coefficient, $\operatorname{RV}(S,C)=\operatorname{tr}(SC)/\sqrt{\operatorname{tr}(S^2)\operatorname{tr}(C^2)}$. As Figure~\ref{fig:patternsImproved} shows, KronSAE retains a stronger correlation structure than TopK SAE and aligns better with the ground-truth covariance matrix. Table~\ref{tab:patternsImproved} shows that KronSAE's learned structure varies more across covariance setups, as reflected by the reported standard deviations, and aligns better with ground truth according to higher RV scores and a smaller effective-rank difference $\Delta$.

\begin{table*}[tb]
\centering
    \begin{tabular}{c|c|c|c|c|c|c}
    \toprule
    Model & RV coefficient	& Effective rank	& Mean correlation	& Rank $\Delta$	\\ \midrule
AE	& 0.59 $\pm$ 0.17	& 63.3 $\pm$ 0.3	& 0.13 $\pm$ 0.04	& 47.9 \\
TopK	& 0.12 $\pm$ 0.02	& 58.5 $\pm$ 1.2	& 0.08 $\pm$ 0.01	& 43.2 \\ 
Kron $h=2,m=4,n=32$	& 0.22 $\pm$ 0.07	& 53.8 $\pm$ 3.4	& 0.08 $\pm$ 0.01	& 38.5 \\
Kron $h=4,m=2,n=32$	& 0.3 $\pm$ 0.12	& 46.0 $\pm$ 7.8	& 0.10 $\pm$ 0.03	& 30.7 \\ 
Kron $h=4,m=8,n=8$	& 0.23 $\pm$ 0.08	& 48.8 $\pm$ 5.0	& 0.09 $\pm$ 0.02	& 33.5	\\
Kron $h=8,m=2,n=16$	& 0.17 $\pm$ 0.04	& 55.9 $\pm$ 1.7	& 0.07 $\pm$ 0.01	& 40.6	\\
    \bottomrule
    \end{tabular}
    \caption{Results with improved matching scheme, computed for 8 different covariance setups. Higher RV coefficients between ground truth and learned matrices and lower effective rank difference $\Delta$ between properties of these matrices indicate that KronSAE is more variable across different setups and is better aligned with ground truth correlation structures.}

    \label{tab:patternsImproved}
\end{table*}

\section{Feature Analysis Methodology} \label{sec:featureAnalysisDetails}

We analyze learned features using an established pipeline \cite{bills2023llmcanexplain,paulo2024automatically} consisting of three stages: (1) statistical property collection, (2) automatic activation pattern interpretation, and (3) interpretation evaluation. 

\subsection{Data collection}

Our collection process uses a fixed-size buffer of $B=384$ examples per feature and continues until reaching a predetermined maximum token count $T_{\max}$. The procedure is as follows.

Initial processing batches generate large activation packs of 1M examples, each comprising a 256-token text segment. When a feature activates, we add the example to its buffer and randomly downsample whenever the buffer exceeds size $B$. This approach supports arbitrary token volumes while accommodating rare features that require extensive sampling.

During collection, we compute online statistics including activation minimums, maximums, means, and frequencies. Post-processing yields two key metrics: token entropy and multitoken ratio. The token entropy is calculated as:
\begin{equation}
\begin{gathered}
    \text{token entropy} = -\sum_{i=1}^{n}p_i \log p_i, \\ p_i = \frac{\text{activations on token } i}{\text{total activations}},
\end{gathered}
\end{equation}
where $n$ is the number of unique activated tokens. The multitoken ratio is computed per sequence:
\begin{equation}
    \text{mt ratio} = \frac{1}{b}\sum_{i=1}^b\frac{\text{activations in example }i}{\text{tokens in example }i},
\end{equation}
with $b < B$ denoting collected context examples per feature.

We then segment examples using a 31-token context window (15 tokens before and after each activation), potentially creating overlapping but non-duplicated examples. Features with a high multitoken ratio may yield significantly more than $B$ context windows.

A separate negative examples buffer captures non-activating contexts. Future enhancements could employ predictive modeling (e.g., using frequent active tokens) to strategically populate this buffer with expected-but-inactive contexts, potentially improving interpretation quality.

\subsection{Feature interpretations}

For each feature, we generate interpretations by sampling 16 activation examples above the median activation and presenting them to Qwen3-14B \citep{yang2025qwen3technicalreport} (AWQ-quantized, with reasoning enabled). The model produces concise descriptions of the activation patterns. Our qualitative observations suggest that reasoning mode improves interpretation quality, although we do not quantify this effect. This is consistent with \citet{paulo2024automatically}, who compare standard responses with chain-of-thought outputs, and motivates further study of reasoning during feature interpretation.

The interpretation process uses the system prompt in Figure~\ref{fig:systemPrompt}. User prompts preserve all special characters because some features activate specifically on them. Figure~\ref{fig:userPrompt} shows a representative, slightly abbreviated user prompt.

\begin{figure}[ht!]
    \centering
\fbox{\begin{minipage}{0.92\linewidth}
{\ttfamily\scriptsize\raggedright
You are a meticulous AI researcher conducting an important investigation into patterns found in language. Your task is to analyze text and provide an explanation that thoroughly encapsulates possible patterns found in it.\par
Guidelines:\par
You will be given a list of text examples in which selected words appear between delimiters like <<this>>. If a sequence of consecutive tokens is important, the entire sequence will appear between delimiters <<just like this>>. The importance of each token is listed after each example in parentheses.\par
- Your explanation should be a concise STANDALONE PHRASE that describes observed patterns.\par
- Focus on the essence of the patterns, concepts, and contexts present in the examples.\par
- Do NOT mention the texts, examples, activations, or the feature itself in your explanation.\par
- Do NOT write ``these texts,'' ``feature detects,'' ``the patterns suggest,'' ``activates,'' or similar phrases.\par
- Do not write what the feature does; e.g., instead of ``detects heart diseases in medical reports,'' write ``heart diseases in medical reports.''\par
- Write the explanation on the last line, exactly after [EXPLANATION]:
}
\end{minipage}}
    \caption{System prompt for feature interpretations.}
    \label{fig:systemPrompt}
\end{figure}

\begin{figure}[ht]
    \centering
\fbox{\begin{minipage}{0.92\linewidth}
{\ttfamily\scriptsize\raggedright
Examples of activations:\par
Text: ' Leno<<,>> a San Francisco Democrat<<, said in a statement.>>'\par
Activations: ' said (22.74), statement (27.84), in (27.54)'\par
Text: ' city spokesman Tyler Gamble<< said in an>> email.'\par
Activations: ' said (2.92), in (12.81), an (14.91)'\par
Text: ' towpath at Brentford Lock. <<Speaking>> on BBC London 94'\par
Activations: 'Speaking (3.48)'\par
Text: ' Michelle, a quadriplegic,<< told>> DrBicuspid.com'\par
Activations: ' told (4.05)'\par
Text: ' CEO Yves Carcelle<< said in a statement>>.'\par
Activations: ' said (19.64), in (29.09), statement (29.39)'
}
\end{minipage}}
    \caption{Example prompt passed to the LLM. This feature received the interpretation ``\textit{Structural elements in discourse, including speech attribution, prepositional phrases, and formal contextual markers},'' with detection and fuzzing scores of 0.84 and 0.76.}
    \label{fig:userPrompt}
\end{figure}

\subsection{Evaluation pipeline} \label{sec:appendix:evaluationPipeline}

We evaluate interpretations using balanced sets of up to 64 positive (activation quantile > 0.5) and 64 negative examples, employing the same model without reasoning to reduce computational costs. When insufficient examples exist, we maintain class balance by equalizing positive and negative counts. The evaluation uses modified system prompts from \citep{paulo2024automatically}, with added emphasis on returning Python lists matching the input example count exactly. We discard entire batches if responses are unparseable or contain fewer labels than the number of provided examples. We calculate two scores.

\textbf{Detection:} After shuffling positive and negative examples, we present up to 8 unformatted text examples per batch to the model. The model predicts an activation label (1/0) for each example, producing up to 128 true/predicted label pairs. We calculate
\begin{equation} \label{eq:score}
    \text{score} = \frac{1}{2}\left(\frac{\text{TP}}{\text{total pos.}} + \frac{\text{TN}}{\text{total negs.}}\right).
\end{equation}
\textbf{Fuzzing:} We $\ll$highlight$\gg$ activated tokens in sampled examples, of which 50\% are correctly labeled positives, 25\% are mislabeled positives, and 25\% are randomly labeled negatives. We present up to 8 examples per batch, and the model evaluates whether each label is correct. Scoring follows Equation~\ref{eq:score}.

\section{Additional Feature Analysis Results} \label{sec:additionalAnalysis}

\begin{figure}[ht]
    \centering
    \includegraphics[width=1.0\linewidth]{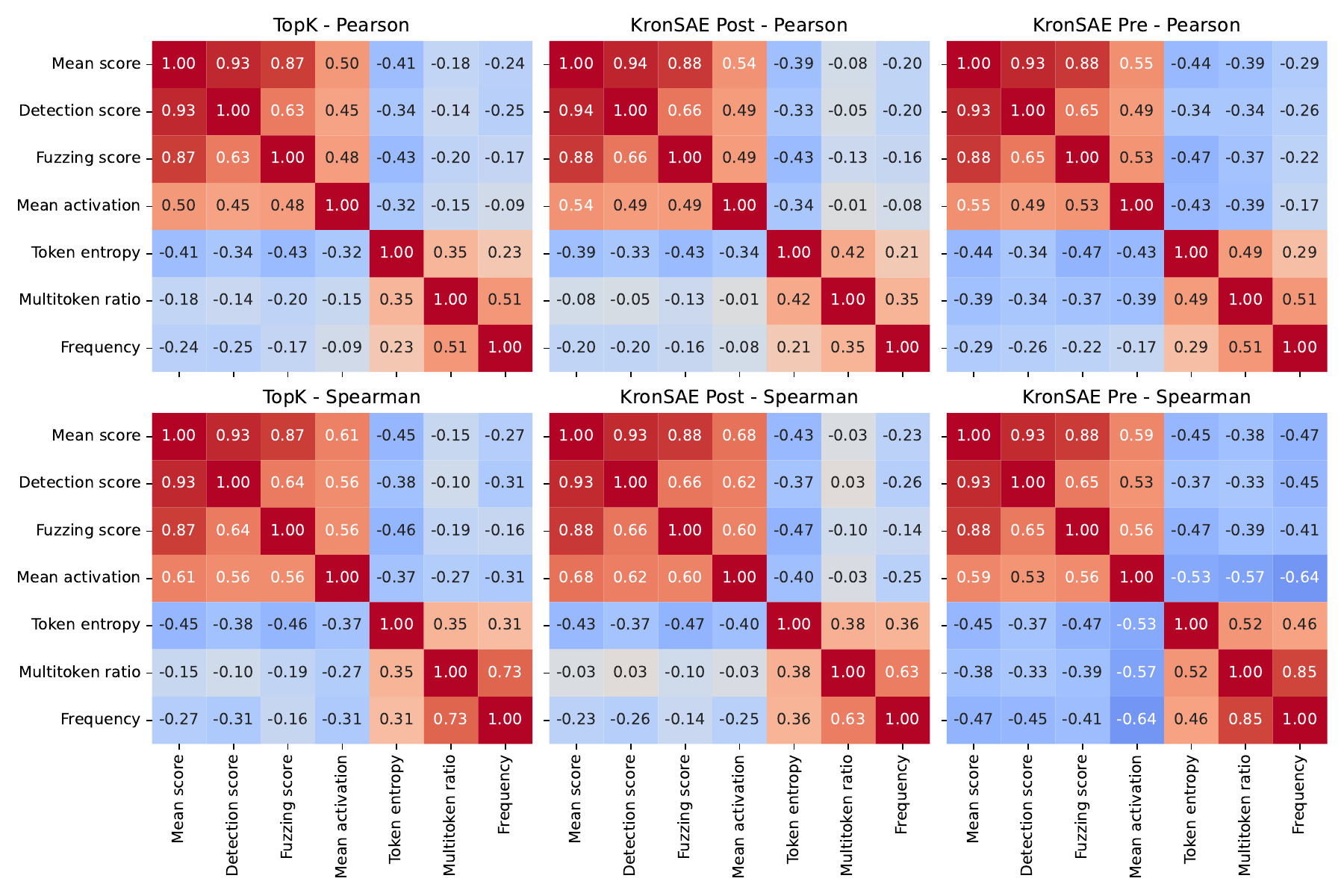}
    \caption{Correlation coefficients (Pearson and Spearman) between properties of TopK and KronSAE latents. Token entropy emerges as a strong predictor of interpretability scores, while higher mean activation and lower frequency also indicate more interpretable features.}
    \label{fig:propertiesCorrelation}
\end{figure}

\paragraph{Feature property correlations.} Our analysis reveals substantial correlations between feature properties and interpretability scores (Figure~\ref{fig:propertiesCorrelation}). In particular, token entropy and mean activation may serve as proxies for feature quality without running the full interpretation pipeline. These findings are based on the first 3,072 features from 32K TopK and KronSAE models ($m=4,n=4$) trained on 24M tokens, and warrant validation at larger scale.

\begin{figure}[ht]
    \centering
    \includegraphics[width=1.0\linewidth]{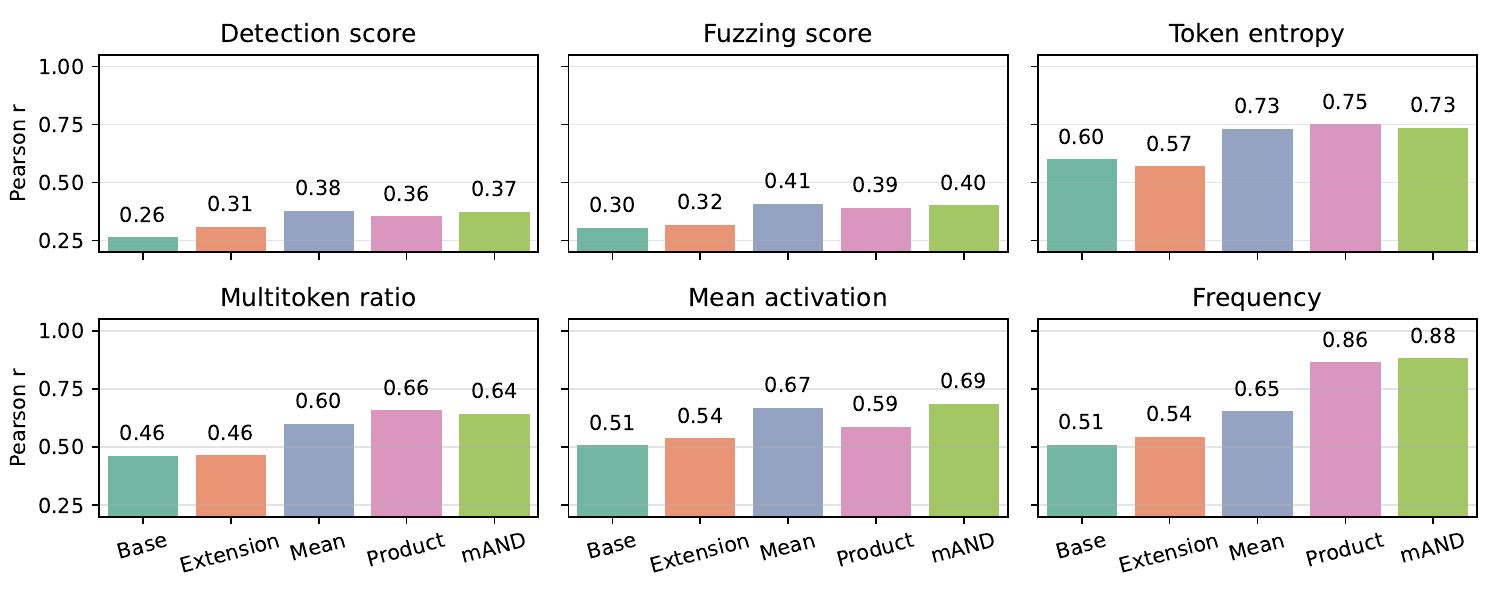}
    \caption{Correlation patterns between properties of post-latents and pre-latents.}
    \label{fig:parentsCorrelations}
\end{figure}

\paragraph{Pre-latent-to-post-latent relationships.} We investigate how post-latent properties correlate with combinations of pre-latent properties, including individual values, means, products, and the $\operatorname{mAND}$ operation (product followed by square root). Figure~\ref{fig:parentsCorrelations} shows that post-latent multitoken ratio, token entropy, and frequency correlate more strongly with pre-latent products or $\operatorname{mAND}$ values than with individual pre-latent properties or their means.

\paragraph{Basis geometry.} Each post-latent has a residual-stream representation given by the corresponding decoder atom, one of the overcomplete basis vectors learned during SAE training. Our architecture clusters feature embeddings: post-latents produced by the same head, base, or extension form tight groups. This geometry depends on the chosen hyperparameters $h$, $m$, and $n$, and may be useful for downstream applications such as steering.

We find that models with more heads achieve better reconstruction while producing more diverse basis vectors. This suggests that fine-grained architectures yield more expressive representations, although they may also exhibit undesired challenging behavior like feature splitting \citep{bricken2023monosemanticity} or absorption \citep{chanin2024absorption}.

\begin{figure}[ht]
    \centering
    \includegraphics[width=1.0\linewidth]{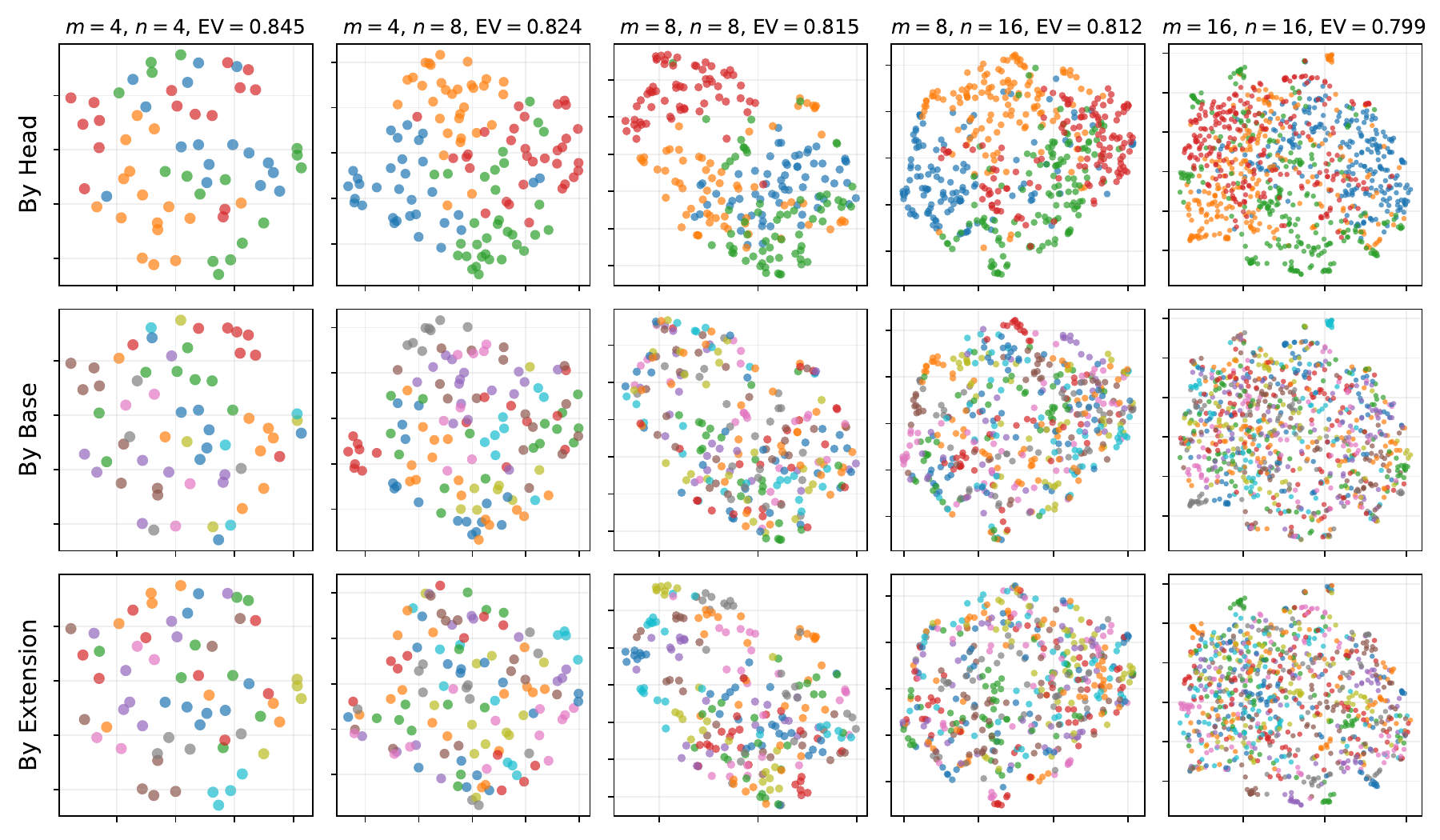}
    \caption{UMAP visualization of post-latent clustering patterns by head, base, and extension group membership. We observe tight clusters by base for $m < n$ and by extension for $m \ge n$.}
    \label{fig:clustering}
\end{figure}

Figure~\ref{fig:clustering} visualizes this structure through UMAP projections (\texttt{n\_neighbors=15}, \texttt{min\_dist=0.05}, and cosine distance) of decoder weights from the first 8 heads of 32K SAEs with varying $m,n$ configurations. For $m<n$, we observe tight base-wise clustering with weaker grouping by extension; for $m\ge n$, extension-wise clustering is stronger.

This asymmetry suggests that pre-latent capacity requirements directly affect the embedding geometry: components with lower polysemanticity (extensions when $m<n$) exhibit greater geometric diversity. We expect symmetric behavior for reciprocal configurations (e.g., $m=4,n=8$ versus $m=8,n=4$), with the roles of bases and extensions exchanged.

\paragraph{Interpretability trade-off.} As Section~\ref{sec:ablations} shows, KronSAE trades off computational efficiency, explained variance, and interpretability. Increasing $m$ and $n$ places more correlated features in each head and improves computational efficiency, but can reduce reconstruction performance. Conversely, smaller $m$ and finer-grained groups improve explained variance at the cost of a slight reduction in interpretability. Lower absorption scores are better; we report them alongside interpretability to make this trade-off explicit. Similar trade-offs occur across a range of SAEs \citep{karvonen2025saebench}.

\section{Analysis of Compositional Structure} \label{sec:compositionalStructure}

\paragraph{Head 3.} 
For this head, we select all base elements and extension 2, as shown in Table~\ref{tab:interactionExampleHead3}. Extension 2 has moderate interpretability and exhibits clear AND-like interactions with base 1 (semantic inheritance through shared pre-latent semantics) and base 2 (retaining only instrument-related semantics). Other interactions occur with base 0 (acquiring medical semantics while preserving metric/number aspects) and base 3 (combining necessity-related semantics with technical or medical concepts to yield therapy/treatment concepts). The latter resembles semantic addition more than intersection. The high interpretability scores suggest encoding mechanisms beyond simple intersection, possibly involving activation magnitude, although dataset or interpretation artifacts cannot be ruled out without further validation.

\begin{table*}[tb]
\centering
\renewcommand{\arraystretch}{1.2}
\begin{tabularx}{\textwidth}{@{} l X c @{}}
\toprule
Component & Interpretation & Score \\
\midrule
\textbf{Extension 2} & Scientific instruments, acronyms, and critical numerical values in technical and astronomical contexts & 0.71 \\
\midrule
\multicolumn{3}{@{}l}{Base elements and their compositions with extension 2} \\
\midrule
\textbf{Base 0} & Punctuation marks and line breaks serving as structural separators in text. & 0.66 \\
\textit{Composition:} & Health-related metrics focusing on survival rates, life expectancy, and longevity. & 0.88 \\
\addlinespace
\textbf{Base 1} & Numerical values, both in digit form and as spelled-out numbers, often accompanied by punctuation like decimals or commas, in contexts of measurements, statistics, or quantitative expressions. & 0.80 \\
\textit{Composition:} & Numerical digits and decimal points within quantitative values. & 0.86 \\
\addlinespace
\textbf{Base 2} & Nuanced actions and adverbial emphasis in descriptive contexts. & 0.71 \\
\textit{Composition:} & Astronomical instruments and their components, such as space telescopes and their acronyms, in scientific and observational contexts. & 0.90 \\
\addlinespace
\textbf{Base 3} & Forms of the verb "to have" indicating possession, necessity, or occurrence in diverse contexts. & 0.91 \\
\textit{Composition:} & Antiretroviral therapy components, viral infection terms, and medical treatment terminology. & 0.87 \\
\bottomrule
\end{tabularx}

\caption{Interactions between extension 2 in head 3 and all base elements in that head.}

\label{tab:interactionExampleHead3}
\end{table*}

\paragraph{Head 136.} 
This head exhibits higher interpretability in post-latents than in pre-latents. Table~\ref{tab:interactionExampleHead136} shows that combining extension 2 with base 0 narrows the semantics to Illinois, likely through a geographical sub-semantic. Interactions with bases 2 and 3 are more complex than simple intersection and introduce additional semantics that require further investigation.

\begin{table*}[tb]
\centering
\renewcommand{\arraystretch}{1.2}

\begin{tabularx}{\textwidth}{@{} l X c @{}}
\toprule
Component & Interpretation & Score \\
\midrule
\textbf{Extension 2} & Hierarchical scopes, geographic references, and spatial dispersal terms & 0.78 \\
\midrule
\multicolumn{3}{@{}l}{Base elements and their compositions with extension 2} \\
\midrule
\textbf{Base 0} & Numerical decimal digits in quantitative expressions and proper nouns. & 0.79 \\
\textit{Composition:} & The state of Illinois in diverse contexts with high significance. & 0.95 \\
\addlinespace
\textbf{Base 1} & The number three and its various representations, including digits, Roman numerals, and related linguistic forms. & 0.84 \\
\textit{Composition:} & Geographic place names and their linguistic variations in textual contexts. & 0.91 \\
\addlinespace
\textbf{Base 2} & Ordinal suffixes and temporal markers in historical or chronological contexts. & 0.87 \\
\textit{Composition:} & Terms indicating layers, degrees, or contexts of existence or operation across scientific, organizational, and conceptual domains. & 0.82 \\
\addlinespace
\textbf{Base 3} & Question formats and topic introductions with specific terms like "What", "is", "of", "the", "Types", "About" in structured text segments. & 0.77 \\
\textit{Composition:} & Spatial spread and occurrence of species or phenomena across environments. & 0.87 \\
\bottomrule
\end{tabularx}
\caption{Interactions between extension 2 in head 136 and all base elements in that head.}
\label{tab:interactionExampleHead136}
\end{table*}

\paragraph{Head 177.} 
The latents in Table~\ref{tab:interactionExampleHead177} demonstrate more consistent AND-like behavior than those in heads 3 and 136, closely matching the interaction pattern in Table~\ref{tab:interactionExample}.

\begin{table*}[tb]
\centering
\renewcommand{\arraystretch}{1.2}

\begin{tabularx}{\textwidth}{@{} l X c @{}}
\toprule
Component & Interpretation & Score \\
\midrule
\textbf{Extension 1} & Geographical mapping terminology and institutional names, phrases involving spatial representation and academic/organizational contexts & 0.90 \\
\midrule
\multicolumn{3}{@{}l}{Base elements and their compositions with extension 1} \\
\midrule
\textbf{Base 0} & Proper nouns, abbreviations, and specific named entities. & 0.64 \\
\textit{Composition:} & Geographical or spatial references using the term "map". & 0.93 \\
\addlinespace
\textbf{Base 1} & Emphasis on terms indicating feasibility and organizations. & 0.80 \\
\textit{Composition:} & Specific organizations and societal contexts. & 0.89 \\
\addlinespace
\textbf{Base 2} & Institutional names and academic organizations, particularly those containing "Institute" or its abbreviations, often paired with prepositions like "of" or "for" to denote specialization or affiliation. & 0.89 \\
\textit{Composition:} & Institutional names containing "Institute" as a core term, often followed by prepositions or additional descriptors. & 0.92 \\
\addlinespace
\textbf{Base 3} & Closure and termination processes, initiating actions. & 0.79 \\
\textit{Composition:} & Initiating or establishing a state, direction, or foundation through action. & 0.85 \\
\bottomrule
\end{tabularx}
\caption{Interactions between extension 1 in head 177 and all base elements in that head.}
\label{tab:interactionExampleHead177}
\end{table*}

\clearpage
\onecolumn
\section{KronSAE Simplified Implementation}
\label{appendix:pseudocode}

\begin{lstlisting}[style=python]
class KronSAE(nn.Module):
    def __init__(self, config):
        super().__init__()
        self.config = config
        _t = torch.nn.init.normal_(
                torch.empty(
                    self.config.act_size, 
                    self.config.h * (self.config.m + self.config.n)
                ) 
            ) / math.sqrt(self.config.dict_size * 2.0)
        self.W_enc = nn.Parameter(_t)
        self.b_enc = nn.Parameter(
            torch.zeros(self.config.h * (self.config.m + self.config.n))
        )
        W_dec_v0 = einops.rearrange( # Initialize decoder weights
            _t.t().clone(), "(h mn) d -> h mn d", 
            h=self.config.h, mn=self.config.m + self.config.n
        )[:, :self.config.m]
        W_dec_v1 = einops.rearrange(
            _t.t().clone(), "(h mn) d -> h mn d", 
            h=self.config.h, mn=self.config.m + self.config.n
        )[:, self.config.m:]
        self.W_dec = nn.Parameter(einops.rearrange(
            W_dec_v0[..., None, :] + W_dec_v1[..., None, :, :], 
            "h m n d -> (h m n) d"
        ))
        self.W_dec.data[:] = (
            self.W_dec.data / self.W_dec.data.norm(dim=-1, keepdim=True)
        )
        self.b_dec = nn.Parameter(torch.zeros(self.config.act_size))

    def encode(self, x: torch.Tensor) -> torch.Tensor:
        B, _ = x.shape
        acts = F.relu(
            x @ self.W_enc + self.b_enc
        ).view(
            B, self.config.h, self.config.m + self.config.n
        )
        base = acts[..., :self.config.m]
        extension = acts[..., self.config.m:]
        all_scores = torch.sqrt(
            base[..., :, None] * extension[..., None, :] + 1e-5
        ).view(B, -1)
        scores, indices = all_scores.topk(
            self.config.k, dim=-1, sorted=False
        )
        acts_topk = torch.zeros_like(all_scores).scatter(
            -1, indices, scores
        )
        return acts_topk

    def forward(self, x):
        acts_topk = self.encode(x)
        x_rec = acts_topk @ self.W_dec + self.b_dec
        output = self.get_loss_dict(x, x_rec)
        return output

    def get_loss_dict(self, x, x_rec):
        loss = (x_rec - x).pow(2).mean()
        var = (x - x.mean(0)).pow(2).mean()
        ev = 1 - loss / var
        return loss, ev

\end{lstlisting}

\end{document}